\pdfoutput=1

\documentclass[11pt]{article}

\usepackage[final]{acl}

\usepackage{times}
\usepackage{latexsym}

\usepackage[T1]{fontenc}

\usepackage[utf8]{inputenc}

\usepackage{microtype}

\usepackage{inconsolata}

\usepackage{graphicx}

\usepackage{microtype}
\usepackage{subfigure}
\usepackage{booktabs} 

\usepackage{amsmath}
\usepackage{amssymb}
\usepackage{mathtools}
\usepackage{amsthm}

\usepackage{url}

\usepackage{colortbl}
\usepackage{multicol}
\usepackage{multirow}

\usepackage{fancybox}
\usepackage{listings}
\usepackage{times}
\usepackage{tabularx}
\usepackage{subcaption}
\usepackage{wrapfig}
\usepackage{algorithm}
\usepackage{algpseudocode}
\usepackage[most]{tcolorbox}
\usepackage{xltabular}
\usepackage{seqsplit}
\usepackage{amsmath,bm}
\usepackage{arydshln}

\definecolor{LightCyan}{rgb}{0.88,1,1}
\definecolor{verylightgray}{rgb}{0.93,0.93,0.93}

\definecolor{kleinblue}{RGB}{0, 47, 167}
\definecolor{lightorange}{RGB}{224, 108, 0}
\definecolor{lightblue}{RGB}{25, 117, 210}
\definecolor{lightgreen}{RGB}{144, 238, 144}

\definecolor{backorange}{RGB}{255, 219, 185}
\definecolor{backblue}{RGB}{202, 225, 249}

\newtcbtheorem[number within=section]{exmp}{Example}%
{beforeafter skip balanced=.01in,colback=white!10!white,colframe=black!50!white,fonttitle=\bfseries, left=.02in, right=.02in,bottom=.02in, top=.02in}{exmp}
\newtcbtheorem[]{prompt}{Prompt}%
{colback=white!10!white,colframe=lightblue!50!white,fonttitle=\bfseries, left=.02in, right=.02in,bottom=.02in, top=.02in}{prompt}

\title{Revisiting LoRA through the Lens of Parameter Redundancy: \\ Spectral Encoding Helps}

\author{
    \textbf{Jiashun Cheng\thanks{Equal contribution.}}, \textbf{Aochuan Chen\footnotemark[1]}, \textbf{Nuo Chen}, \textbf{Ziqi Gao}, \textbf{Yuhan Li}, \\
    \textbf{Jia Li\thanks{Corresponding author.}}, \textbf{Fugee Tsung} \\
    \\
    The Hong Kong University of Science and Technology (Guangzhou) \\
    The Hong Kong University of Science and Technology \\
    \texttt{jchengak@connect.ust.hk, jialee@ust.hk}
}

\begin{document}
\maketitle
\begin{abstract}
Low-Rank Adaptation (LoRA) has emerged as a prominent technique for fine-tuning large foundation models.
Despite its successes, the substantial parameter redundancy, which limits the capacity and efficiency of LoRA, has been recognized as a bottleneck.
In this work, we systematically investigate the impact of redundancy in fine-tuning LoRA and reveal that reducing density redundancy does not degrade expressiveness.
Based on this insight, we introduce \underline{S}pectral-\underline{e}ncoding \underline{L}ow-\underline{R}ank \underline{A}daptation (SeLoRA), which harnesses the robust expressiveness of spectral bases to re-parameterize LoRA from a sparse spectral subspace. 
Designed with simplicity, SeLoRA enables seamless integration with various LoRA variants for performance boosting, serving as a scalable plug-and-play framework.
Extensive experiments substantiate that SeLoRA achieves greater efficiency with fewer parameters, delivering superior performance enhancements over strong baselines on various downstream tasks, including commonsense reasoning, math reasoning, and code generation.
\end{abstract}

\section{Introduction}\label{sec:intro}

In recent years, Large Foundation Models (LFMs), have showcased exceptional generalization capabilities, greatly improving performance in a wide array of tasks across natural language processing (NLP)~\citep{brown2020language, touvron2023llama}, computer vision (CV)~\citep{radford2021learning, kirillov2023segment}, and other fields~\citep{azad2023foundational, li2024glbench, li2025g}. 
Typically, adapting these general models for specific downstream tasks requires full fine-tuning, which involves retraining all model parameters and can pose significant challenges, particularly in resource-limited environments.
To address this issue, Parameter-efficient fine-tuning (PEFT) techniques~\citep{peft}, have been developed, offering more feasible alternatives.
Among these, Low-Rank Adaptation (LoRA)~\citep{hu2021lora}, which decomposes the weight changes into the product of two low-rank matrices, has stood out for its effectiveness and simplicity.

\begin{figure}[t]
    \includegraphics[width=\linewidth]{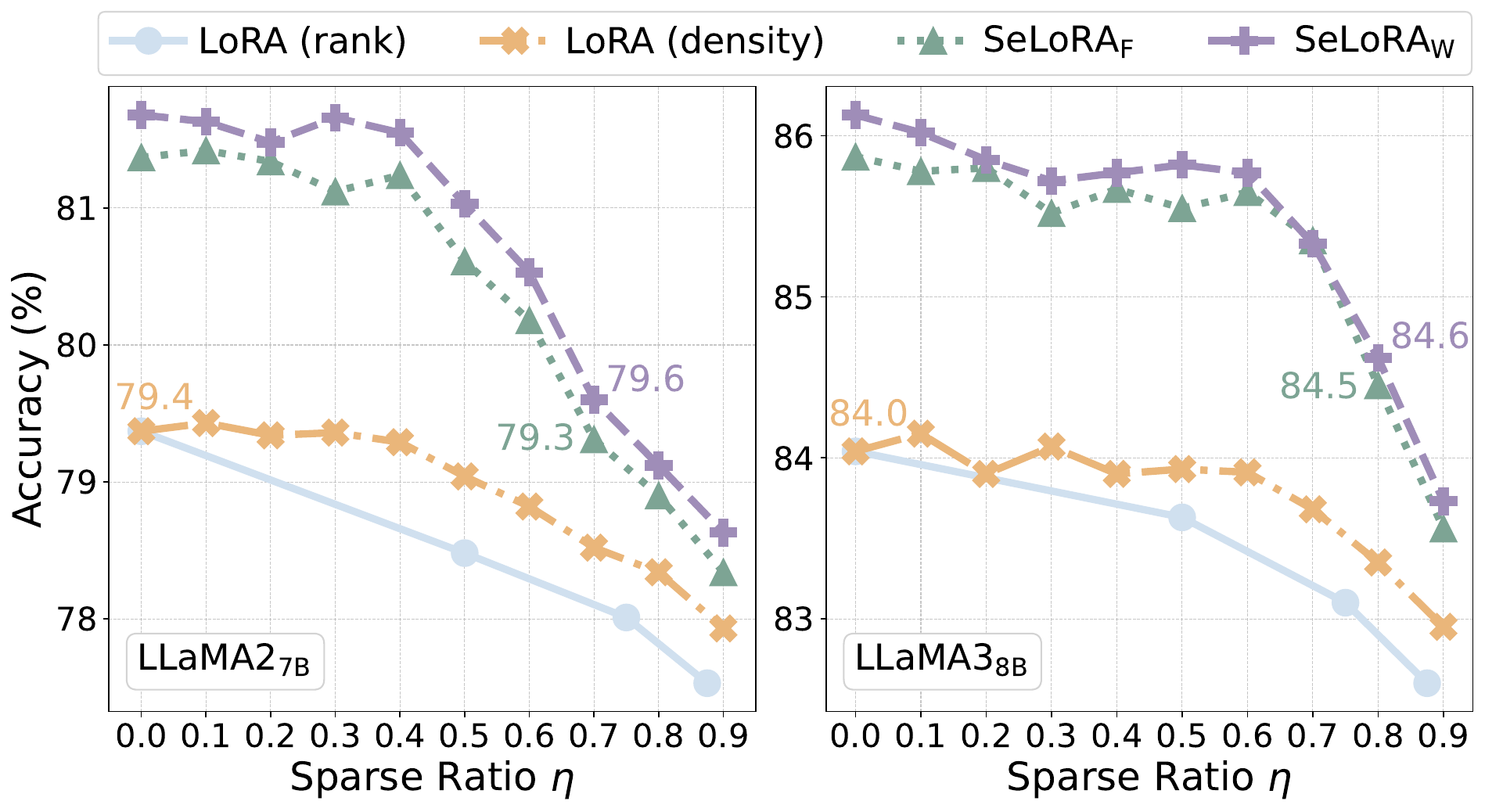}
    \caption{
    Average accuracy of commonsense reasoning tasks (\textbf{y-axis}) across various sparse ratios $\eta$ (\textbf{x-axis}) on trainable parameters.
    Masking a significant proportion of LoRA's parameters still retains comparable performance, while naively learning with reduced ranks leads to clear degradation. 
    }
    \label{fig:motivation}
\end{figure}

Despite its success, recent studies still indicate redundancy in LoRA’s parameters.
Early work~\citep{he2022sparseadapter} conducts a systematic exploration into such redundancy in encoder-only models.
For decoder-only LFMs, subsequent approaches~\citep{kopiczko2023vera, renduchintala2023tied} reduce the parameters by sharing frozen LoRA matrices across layers and modules while learning small scaling vectors with only slight performance losses.
More recent efforts~\citep{balazy2024lora, li2024vblora, sehanobish2024structured} further minimized the parameter usage by re-parameterizing LoRA via matrix decomposition.
Concurrently, another line of research~\citep{ding2023sparse, jiang2024taia, jiang2025finetuning} unveils that removing the redundant components from the fine-tuned LoRA's parameters can yield further performance gains.
However, despite these advancements, a systematic understanding of how redundancy affects LoRA in the context of decoder-only LFMs during the fine-tuning phase remains absent.

To address this gap, drawing inspiration from prior works~\citep{he2022sparseadapter, yu2024language}, we conduct a comprehensive investigation into the impacts of redundancy in LoRA from two perspectives: 
(1) \textbf{Rank redundancy} - Fine-tuning LoRA with reduced ranks; 
(2) \textbf{Density redundancy} - Masking a portion of LoRA's parameters to zero while fine-tuning the remainder at fixed rank.
As presented in Figure~\ref{fig:motivation}, we observe that reducing the rank alone leads to notable performance degradation.
Conversely, when fine-tuning LoRA at an appropriately chosen rank, the introduction of sparsity, achieved by masking up to 60\% of the parameters, demonstrates performance on par with that of the fully parameterized LoRA.
It highlights that adequately reducing density redundancy does not compromise its expressiveness and we term this phenomenon the \textit{sparsity property} of LoRA.
These findings suggest that LoRA's parameters are not fully utilized, leaving room for further enhancement.
In light of these observations, a question is naturally raised:

\begin{center}
\textit{How can we unleash the potential of LoRA \\ utilizing its sparsity property?}
\end{center}

This question aligns closely with the principles of sparse learning~\citep{han2015deep}, which seeks to acquire expressive information while necessitating fewer learnable parameters. 
Despite the success of the predominant pruning techniques~\citep{han2015learning, frankle2018lottery}, recent studies~\citep{zhao2024apt, gu2024light} demonstrate effective pruning during LoRA fine-tuning often requires intricate strategies.
In contrast, spectral encoding of weight matrices~\citep{koutnik2010evolving, van2016wavelet}, which enables expressive representation learning with sparse spectral entries~\citep{wolter2020neural, irie2021training}, offers a more straightforward yet powerful alternative.

Motivated by these findings, we introduce \underline{S}pectral-\underline{e}ncoding \underline{L}ow-\underline{R}ank \underline{A}daptation (SeLoRA), a novel approach that harness spectral transformations to re-parameterize low-rank matrices as the spatial equivalents of spectral components.
Essentially, SeLoRA selectively learns only a sparse set of spectral components at the predefined globally shared spectral locations, where inverse spectral transformation is then applied to derive the adaptation matrices in the spatial domain.
Designed with simplicity and flexibility, SeLoRA naturally accommodates various choices of spectral bases, making it highly flexible.
Furthermore, its lightweight and modular nature enables seamless integration as a plug-and-play framework compatible with a variety of LoRA variants.
Our evaluation of SeLoRA on advanced LFMs like LLaMA families across diverse instruction-tuning tasks demonstrates its enhanced capacity and efficiency, as exemplified in Figure~\ref{fig:motivation}.
Extensive in-depth analyses are further conducted to substantiate the robustness of SeLoRA, confirming its advantages and practical utility.

In summary, our contributions are as follows:
\begin{itemize}
    \item Our investigation into the impact of redundancy in LoRA reveals the \textit{sparsity property}, where optimizing only a sparse subset of tunable parameters preserves comparable expressive power.
    \item Based on this insight, we introduce SeLoRA, a novel extension of LoRA with expressive spectral re-parameterization, enhancing performance while reducing parameter overhead. 
    \item We rigorously evaluate SeLoRA across multiple domains, validating its effectiveness and efficiency. 
    A comprehensive analysis further elucidates the impact of its designs.
\end{itemize}
\section{Methodology}\label{sec:method}

\begin{figure*}[t]
    \centering
    \vspace{-5mm}
    \includegraphics[width=1.0\textwidth]{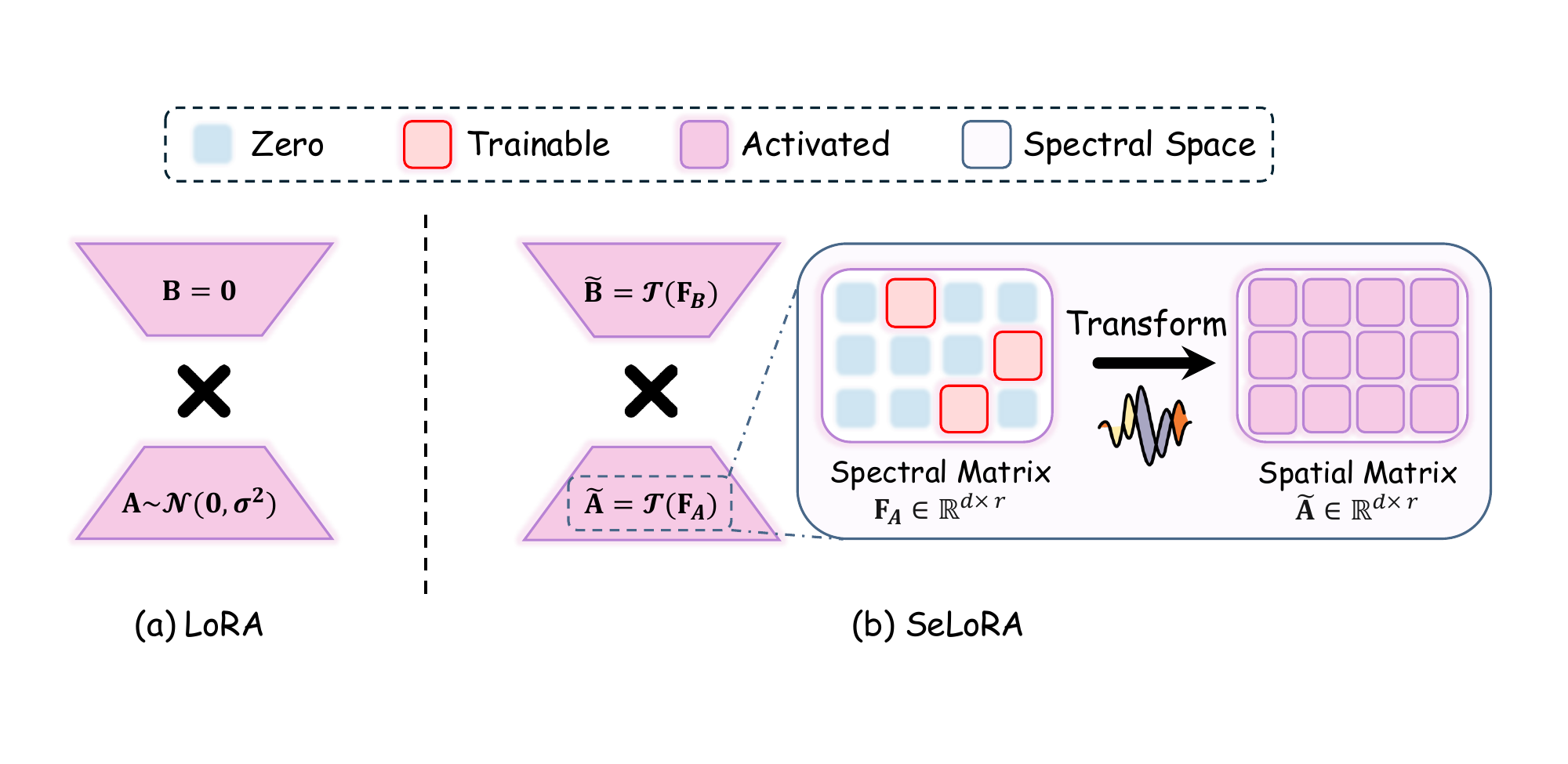}
    \vspace{-5mm}
    \caption{
    An overview of the schematic comparison between LoRA and our proposed SeLoRA.
    In contrast to fully parameterized LoRA, SeLoRA employs spectral transformations on sparse spectral components to represent weight matrices. 
    } 
    \label{fig:main}
\end{figure*}

In this section, we first outline the basic properties of LoRA fine-tuning. 
We then present our proposed \underline{S}pectral-\underline{e}ncoding \underline{L}ow-\underline{R}ank \underline{A}daptation (SeLoRA), which takes advantage of the \textit{sparsity property} along with spectral transformations for effective representation learning.
The overall framework is presented in Figure~\ref{fig:main}.

\subsection{Background}\label{sec:background}

\paragraph{Low-Rank Adaptation.}
LoRA~\citep{hu2021lora} assumes parameter changes typically occur within a low-rank space~\citep{aghajanyan2020intrinsic} and proposes to use the product of two low-rank matrices $\mathbf{B} \in \mathbb{R}^{d_{1} \times r}$ and $\mathbf{A} \in \mathbb{R}^{r \times d_{2}}$ as the incremental weight update $\Delta \mathbf{W} \in \mathbb{R}^{d_{1} \times d_{2}}$.
For pre-trained weight $\mathbf{W}_{0} \in \mathbb{R}^{d_{1} \times d_{2}}$, LoRA is expressed as
\begin{equation}
   \mathbf{W}' = \mathbf{W}_{0} + \Delta \mathbf{W} = \mathbf{W}_{0} + \mathbf{B} \mathbf{A},
\end{equation}
where $\mathbf{A}, \mathbf{B}$ are trainable with the rank $r \ll \{ d_{1}, d_{2} \}$ while $\mathbf{W}_{0}$ is frozen during fine-tuning. 
Without loss of generality, we assume $d = d_{1} = d_{2}$ for notation-wise simplicity.

\subsection{LoRA with Spectral Encoding}

As previously discussed, our objective is to enhance the expressiveness of $\mathbf{A}$ and $\mathbf{B}$ while learning with reduced density redundancy, which aligns closely with the foundational principle of sparse learning.
Upon revisiting prior successes, we adopt the spectral encoding for weight matrices, a well-established approach that balances simplicity and expressiveness~\citep{wolter2020neural, irie2021training}.

Essentially, our approach centers on re-parameterizing the adaptation matrices, termed $\tilde{\mathbf{A}} \in \mathbb{R}^{r \times d}$ and $\tilde{\mathbf{B}} \in \mathbb{R}^{d \times r}$, as the spatial recovery of sparse spectral components, while retaining LoRA's update schema:
\begin{equation}\label{eqn:fora_full}
    \mathbf{W}' = \mathbf{W}_{0} + \Delta \mathbf{W} = \mathbf{W}_{0} + \tilde{\mathbf{B}} \tilde{\mathbf{A}}.
\end{equation}

To achieve this, we first introduce the sparse ratio $\eta \in (0, 1)$, representing the proportion of masking parameters relative to the total elements in the low-rank matrix. 
We then randomly initialize an index set $\Omega$, where $|\Omega| = \lfloor (1 - \eta) \cdot rd \rfloor$, specifying the locations of the learnable spectral components shared across the low-rank matrices.
Subsequently, we define the corresponding sparse spectral matrices as $\mathbf{F}_{A} \in \mathbb{R}^{r \times d}$ and $\mathbf{F}_{B} \in \mathbb{R}^{d \times r}$.
For clarity, we denote $\mathbf{F}_{A}(u, v)$ as the element at index $(u, v)$.
The matrix $\mathbf{F}_{A}$ is constructed such that its entries $\mathbf{F}_{A}(u, v)$ are learnable if $(u, v) \in \Omega$, while all other entries are fixed to zero.
The same principle applies to the construction of $\mathbf{F}_{B}$.
Given such, the low-rank matrices are derived from their spatial counterpart by applying various inverse spectral transformations:
\begin{equation}\label{eqn:spctral_trans}
    \tilde{\mathbf{A}} = \mathcal{T}(\mathbf{F}_{A}), \, \tilde{\mathbf{B}} = \mathcal{T}(\mathbf{F}_{B}),
\end{equation}
where $\mathcal{T}(\cdot)$ represents transformation function.

As demonstrated, the core of this framework is leveraging efficient and expressive spectral bases to enhance learning capacity.
To this end, we explore two widely adopted and well-established spectral transformations as representative instantiations of our approach. 
We exemplify these two variants via the computation of $\tilde{\mathbf{A}}$, and $\tilde{\mathbf{B}}$ is computed by applying the identical procedure.

\paragraph{Fourier Encoding.}

Known for the capacity to capture high-fidelity information from sparse spectral components~\citep{duarte2013spectral, vlaardingerbroek2013magnetic}, the Fourier basis emerges as an indispensable choice.
In light of that, we employ the discrete inverse 2D Fourier transform, denoted as $\mathcal{F}^{-1}(\cdot)$, and retain only the real part of the transformed results to simplify computations.
Accordingly, the resulting method is defined as the following:
\begin{equation}\label{eqn:idft}
{
\begin{aligned}
    \tilde{\mathbf{A}}(j, k) & = \text{Re}[ \mathcal{F}^{-1} (\mathbf{F}_{A}) ] \\ 
    & = \text{Re} \big[ \sum_{u, v} \mathbf{F}_{A}(u, v) \, e^{i 2\pi (\frac{u}{r} j + \frac{v}{d} k)} \big],
\end{aligned}
}
\end{equation}
where $i$ denotes the imaginary unit and the transformation function is formulated as $\mathcal{T}(\cdot) = \text{Re}[\mathcal{F}^{-1}(\cdot)]$.

\paragraph{Wavelet Encoding.}

Upon revisiting prior successes, the Wavelet basis also presents as a promising choice~\cite {van2016wavelet}.
Unlike the Fourier transform, which primarily captures global frequency information, the Wavelet transform provides a more localized and hierarchical reconstruction of information with flexible filter options. 
In this paper, we adopt the discrete inverse 2D Wavelet transform, denoted as  $\mathcal{T}(\cdot) = \mathcal{W}^{-1}(\cdot)$, that is widely utilized in image restoration.

Following the standard practice, we first decompose the spectral matrix $\mathbf{F}_{A}$ into four spectral components:
\begin{equation}\label{eqn:wavelet_split}
    \mathbf{F}_{A} = 
    \left[
    \begin{array}{cc}
        \mathbf{F}_{A}^{a} & \mathbf{F}_{A}^{h} \\
        \mathbf{F}_{A}^{v} & \mathbf{F}_{A}^{d} \\
    \end{array}
    \right], 
\end{equation}
where $\mathbf{F}_{A}^{a} \in \mathbb{R}^{\frac{r}{2} \times \frac{d}{2}}$ denotes the approximation coefficients, while the remaining $\mathbf{F}_{A}^{o} \in \mathbb{R}^{\frac{r}{2} \times \frac{d}{2}}, \, o \in \{h, v, d\}$ correspond to detail coefficients of different directions. 
Let $O = \{a , h, v, d\}$ denote the set of coefficient types, the transformation is then expressed as:
\begin{equation}\label{eqn:idwt}
{
\begin{aligned}
    \tilde{\mathbf{A}}(j, k) & = \mathcal{W}^{-1} (\mathbf{F}_{A}) \\ 
    & = \sum_{o \in O} \sum_{u, v} \mathbf{F}_{A}^{o}(u, v) \, \psi^{o}_{u, v}(j, k),
\end{aligned}
}
\end{equation}
with
\begin{equation}
    \psi^{o}_{u, v}(j, k) = \frac{1}{\sqrt{2}} \psi^{o}(\frac{j}{2} - u, \frac{k}{2} - v),
\end{equation}
where $\psi^{o}(\cdot, \cdot)$ denotes the basis function of wavelet filters. 
For instance, when employing the Haar wavelet, $\psi^{o}(\cdot, \cdot)$ reduces to an indicator function taking values in $\{ -1, 1 \}$, active within the range $2u \le j \le 2u+1, \, 2v \le k \le 2v + 1$.
The choice of wavelet filter directly influences the balance between smoothness and detail preservation in the transformed representation.
The specific construction details of wavelet filters are provided in the Appendix~\ref{appedix:wavelet}.
Unless otherwise specified, we use the Haar wavelet as the default basis in this study.

\subsection{Discussions}

\paragraph{Initialization Strategies.}
Matrix initialization with consistent variance~\citep{glorot2010understanding} is crucial for maintaining numerical stability and accelerating convergence.
However, unlike LoRA, directly initializing the spectral space in SeLoRA can lead to suboptimal variance in spatial space due to the involvement of spectral transformations.
For matrix $\tilde{\mathbf{A}}$, we first employ Xavier~\citep{glorot2010understanding} or Kaiming initialization~\citep{he2015delving} to the spectral matrix $\mathbf{F}_{A}$ and an auxiliary matrix $\mathbf{A}' \in \mathbb{R}^{r \times d}$.
Next, we scale $\mathbf{F}_{A}$ to ensure $\text{Var}(\tilde{\mathbf{A}}) = \text{Var}(\mathcal{T}(\mathbf{F}_{A})) = \text{Var}(\mathbf{A}')$.
In contrast, matrix $\tilde{\mathbf{B}}$ is initialized to zeros following the standard practice of LoRA~\citep{hu2021lora}.
We employ Kaiming initialization by default unless specially stated.

\paragraph{Extension to LoRA Variants.}
Unlike introducing an entirely new learning paradigm, our method capitalizes on the modular nature of spectral encoding for weight matrices, enabling seamless integration as a plug-in within various LoRA variants, including DoRA~\citep{liu2024dora}, X-LoRA~\citep{buehler2024x} and HiRA~\citep{huang2025hira}.
Moreover, by leveraging fast spectral transformation~\citep{nussbaumer1982fast, wolter2024ptwt}, our approach introduces only minimal additional computational cost during training while incurring no extra overhead during inference, making it a highly efficient and scalable solution.
A more detailed empirical evaluation is provided in Section~\ref{sec:experiment}.
\section{Experiments}\label{sec:experiment}

\subsection{Setup}\label{sec:setup}

\begin{table*}[htb]
    \centering
    \resizebox{1.\textwidth}{!}{
        \begin{tabular}{lr*{10}{c}}
            \toprule
            Methods & Params (\%) & Time & BoolQ & PIQA & SIQA & HellaS.&  WinoG. & ARC-e & ARC-c & OBQA & Avg. \\
            \midrule \specialrule{0em}{1.5pt}{1.5pt}
            \multicolumn{1}{c}{\textbf{\textit{GPT-3.5-turbo}}} \\
            Zero-shot & - & - & 73.1 & 85.4 & 68.5 & 78.5 & 66.1 & 89.8 & 79.9 & 74.8 & 77.0 \\
            
            \midrule[0.6pt]
            \midrule[0.6pt]

            \multicolumn{1}{c}{\textbf{\textit{LLaMA2$_{\textsc{7b}}$}}} \\

            $\mathcal{LS}$-LoRA & 0.50 & 7.7h & 72.2 & 82.6 & 80.2 & 89.7 & 83.2 & 84.2 & 69.6 & 82.7 & 80.6 \\
            LoRETTA$_{rep}$ & 0.53 & 8.5h & 71.9 & 82.4 & 80.0 & 90.1 & 83.3 & 84.4 & 69.1 & 82.3 & 80.4 \\
            \cmidrule(lr){1-1} \cmidrule(lr){2-12}
            
            LoRA & 0.83 & 7.4h & 71.4 & 81.4 & 79.6 & 87.8 & 83.2 & 82.6 & 67.5 & 81.5 & 79.4 \\
            \textbf{SeLoRA$_{\text{F}}$} & 0.50 & 7.6h & 72.8 & \colorbox{backblue}{83.4} & 80.0 & 90.9 & \colorbox{backblue}{83.7} & \colorbox{backblue}{85.4} & 70.5 & \colorbox{backblue}{\textbf{83.4}} & 81.3 {\scriptsize (\textcolor{blue}{$\uparrow$ 1.9})} \\
            \textbf{SeLoRA$_{\text{W}}$} & 0.50 & 7.5h & \colorbox{backblue}{72.9} & 83.3 & \colorbox{backblue}{80.5} & \colorbox{backblue}{\textbf{92.1}} & 83.5 & 85.3 & \colorbox{backblue}{\textbf{71.9}} & 83.2 & \colorbox{backblue}{81.6 {\scriptsize (\textcolor{blue}{$\uparrow$ 2.2})}} \\
            \cmidrule(lr){1-1} \cmidrule(lr){2-12}
            
            DoRA & 0.84 & 12.2h & 71.8 & 83.1 & 77.1 & 90.1 & 82.8 & 84.1 & 69.5 & 82.4 & 80.1 \\
            \textbf{SeDoRA$_{\text{F}}$} & 0.51 & 12.7h & 72.5 & 83.4 & 80.2 & 91.1 & 84.2 & 85.3 & \colorbox{backblue}{71.7} & \colorbox{backblue}{83.2} & 81.5 {\scriptsize (\textcolor{blue}{$\uparrow$ 1.4})} \\
            \textbf{SeDoRA$_{\text{W}}$} & 0.51 & 12.4h & \colorbox{backblue}{\textbf{73.7}} & \colorbox{backblue}{83.8} & \colorbox{backblue}{\textbf{80.6}} & \colorbox{backblue}{92.0} & \colorbox{backblue}{\textbf{84.6}} & \colorbox{backblue}{\textbf{86.0}} & 71.6 & 83.0 & \colorbox{backblue}{\textbf{81.9} {\scriptsize (\textcolor{blue}{$\uparrow$ \textbf{1.8}})}} \\
            \cmidrule(lr){1-1} \cmidrule(lr){2-12}
            
            HiRA & 0.83 & 11.7h & 72.0 & 82.1 & 78.7 & 86.7 & 79.6 & 84.3 & 70.1 & 79.6 & 79.1 \\
            \textbf{SeHiRA$_{\text{F}}$} & 0.50 & 12.1h & 71.8 & 83.1 & 79.3 & 89.9 & 81.2 & 84.5 & 70.0 & 80.8 & 80.1 {\scriptsize (\textcolor{blue}{$\uparrow$ 1.0})} \\
            \textbf{SeHiRA$_{\text{W}}$} & 0.50 & 11.9h & \colorbox{backblue}{73.2} & \colorbox{backblue}{\textbf{84.2}} & \colorbox{backblue}{79.9} & \colorbox{backblue}{90.8} & \colorbox{backblue}{83.2} & \colorbox{backblue}{\textbf{86.0}} & \colorbox{backblue}{70.8} & \colorbox{backblue}{81.6} & \colorbox{backblue}{81.2 {\scriptsize (\textcolor{blue}{$\uparrow$ 2.1})}} \\
            
            \midrule[0.6pt]
            \midrule[0.6pt]
            
            \multicolumn{1}{c}{\textbf{\textit{LLaMA3$_{\textsc{8b}}$}}} \\
            
            $\mathcal{LS}$-LoRA & 0.28 & 8.1h & 74.2 & 88.0 & 79.6 & 94.7 & 85.2 & 90.1 & 79.1 & 85.4 & 84.5 \\
            LoRETTA$_{rep}$ & 0.30 & 9.0h & 74.5 & 87.8 & 79.7 & 94.6 & 85.4 & 89.7 & 78.2 & 87.0 & 84.6 \\
            \cmidrule(lr){1-1} \cmidrule(lr){2-12}
            
            LoRA & 0.70 & 7.8h & 74.0 & 88.2 & 80.4 & 94.0 & 85.5 & 87.5 & 78.1 & 84.0 & 84.0 \\
            \textbf{SeLoRA$_{\text{F}}$} & 0.28 & 8.1h & 74.4 & 89.0 & \colorbox{backblue}{81.3} & 95.6 & \colorbox{backblue}{\textbf{87.5}} & 90.6 & 80.3 & \colorbox{backblue}{87.0} & 85.7 {\scriptsize (\textcolor{blue}{$\uparrow$ 1.7})} \\
            \textbf{SeLoRA$_{\text{W}}$} & 0.28 & 8.0h & \colorbox{backblue}{76.0} & \colorbox{backblue}{89.3} & 80.6 & \colorbox{backblue}{95.9} & 86.7 & \colorbox{backblue}{91.0} & \colorbox{backblue}{81.4} & 86.6 & \colorbox{backblue}{85.9 {\scriptsize (\textcolor{blue}{$\uparrow$ 1.9})}} \\
            \cmidrule(lr){1-1} \cmidrule(lr){2-12}
            
            DoRA & 0.71 & 12.8h & 74.9 & 88.9 & 80.2 & 95.5 & 85.6 & 90.5 & 80.4 & 85.8 & 85.2 \\
            \textbf{SeDoRA$_{\text{F}}$} & 0.28 & 13.3h & 75.7 & 89.3 & 81.2 & 95.8 & 86.9 & \colorbox{backblue}{91.8} & 81.2 & 86.8 & 86.1 {\scriptsize (\textcolor{blue}{$\uparrow$ 0.9})} \\
            \textbf{SeDoRA$_{\text{W}}$} & 0.28 & 13.0h & \colorbox{backblue}{\textbf{76.2}} & \colorbox{backblue}{\textbf{89.7}} & \colorbox{backblue}{\textbf{81.4}} & \colorbox{backblue}{\textbf{96.0}} & \colorbox{backblue}{\textbf{87.5}} & 91.2 & \colorbox{backblue}{\textbf{82.0}} & \colorbox{backblue}{87.8} & \colorbox{backblue}{\textbf{86.5} {\scriptsize (\textcolor{blue}{$\uparrow$ \textbf{1.3}})}} \\
            \cmidrule(lr){1-1} \cmidrule(lr){2-12}

            HiRA & 0.70 & 12.3h & 74.6 & 88.3 & 79.7 & 95.3 & 85.3 & 90.8 & 80.3 & 88.0 & 85.3 \\
            \textbf{SeHiRA$_{\text{F}}$} & 0.28 & 12.8h & 75.7 & \colorbox{backblue}{89.4} & \colorbox{backblue}{81.1} & 95.5 & 86.5 & 91.6 & 80.8 & 87.0 & 86.0 {\scriptsize (\textcolor{blue}{$\uparrow$ 0.7})} \\
            \textbf{SeHiRA$_{\text{W}}$} & 0.28 & 12.5h & \colorbox{backblue}{76.0} & 89.1 & 80.8 & \colorbox{backblue}{95.9} & \colorbox{backblue}{87.3} & \colorbox{backblue}{\textbf{92.2}} & \colorbox{backblue}{81.8} & \colorbox{backblue}{\textbf{88.8}} & \colorbox{backblue}{\textbf{86.5} {\scriptsize (\textcolor{blue}{$\uparrow$ \textbf{1.2}})}} \\
            
            \bottomrule
        \end{tabular}
    }
    \caption{Comparison of LLaMA2$_{\textsc{7b}}$ and LLaMA3$_{\textsc{8b}}$ against various methods on eight commonsense datasets.
    The results of all baseline methods are reproduced by implementing their official codebase.
    The highest scores within each baseline method are highlighted in \colorbox{backblue}{blue} and the best results of each LLM are marked in \textbf{bold}. 
    Training time is measured on 1x A100 GPU.
    }
    \label{tab:common}
\end{table*}
\begin{table*}[htb]
    \centering
    \resizebox{1.\textwidth}{!}{
        \begin{tabular}{lr*{9}{c}}
            \toprule
            & & & \multicolumn{2}{c}{Math} & & \multicolumn{4}{c}{Code} & \\
            \cmidrule(lr){4-5} \cmidrule(lr){7-10}
            Methods & Params (\%) & GPU (GB) & GSM8k & MATH & Avg. & HumanEval & HumanEval+ & MBPP & MBPP+ & Avg. \\
            \midrule \specialrule{0em}{1.5pt}{1.5pt}
            
            \multicolumn{1}{c}{\textbf{\textit{LLaMA2$_{\textsc{7b}}$}}} \\
            Zero-shot & - & - & 7.3 & 1.1 & 4.2 & 11.0 & 9.8 & 30.2 & 24.1 & 18.8\\
            FourierFT & 0.66 & 47.8 & 61.5 & 10.9 & 36.4 & 31.6 & 27.6 & 37.4 & 32.4 & 32.2 \\
            \cmidrule(lr){1-1} \cmidrule(lr){2-11}
            
            LoRA & 0.83 & 42.2 & 60.5 & 11.7 & 36.1 & 32.1 & 28.4 & 35.8 & 30.8 & 31.8 \\
            \textbf{SeLoRA$_{\text{F}}$} & 0.66 & 42.9 & 61.4 & 12.5 & 37.0 {\scriptsize (\textcolor{blue}{$\uparrow$ 0.9})} & 31.2 & 26.9 & 38.9 & 32.5 & 32.4 {\scriptsize (\textcolor{blue}{$\uparrow$ 0.6})} \\
            \textbf{SeLoRA$_{\text{W}}$} & 0.66 & 42.9 & \colorbox{backblue}{62.4} & \colorbox{backblue}{13.7} & \colorbox{backblue}{38.1 {\scriptsize (\textcolor{blue}{$\uparrow$ 2.0})}} & \colorbox{backblue}{\textbf{35.2}} & \colorbox{backblue}{29.1} & \colorbox{backblue}{40.1} & \colorbox{backblue}{34.8} & \colorbox{backblue}{\textbf{34.8} {\scriptsize (\textcolor{blue}{$\uparrow$ \textbf{3.0}})}} \\
            \cmidrule(lr){1-1} \cmidrule(lr){2-11}
            
            DoRA & 0.84 & 56.4 & 61.2 & 12.1 & 36.7 & 32.9 & 28.7 & 39.9 & 33.1 & 33.7 \\
            \textbf{SeDoRA$_{\text{F}}$} & 0.67 & 57.1 & 62.0 & 12.8 & 37.4 {\scriptsize (\textcolor{blue}{$\uparrow$ 0.9})} & 31.5 & 27.8 & \colorbox{backblue}{\textbf{41.6}} & \colorbox{backblue}{\textbf{35.8}} & 34.2 {\scriptsize (\textcolor{blue}{$\uparrow$ 0.5})} \\
            \textbf{SeDoRA$_{\text{W}}$} & 0.67 & 57.0 & \colorbox{backblue}{\textbf{63.0}} & \colorbox{backblue}{\textbf{14.1}} & \colorbox{backblue}{\textbf{38.6} {\scriptsize (\textcolor{blue}{$\uparrow$ \textbf{1.9}})}} & \colorbox{backblue}{33.5} & \colorbox{backblue}{\textbf{29.9}} & 41.0 & 34.7 & \colorbox{backblue}{\textbf{34.8} {\scriptsize (\textcolor{blue}{$\uparrow$ \textbf{1.1}})}} \\
            
            \midrule[0.6pt]
            \midrule[0.6pt]

            \multicolumn{1}{c}{\textbf{\textit{LLaMA3$_{\textsc{8b}}$}}} \\
            Zero-shot & - & - & 33.1 & 5.3 & 19.2 & 33.5 & 29.3 & 61.4 & 51.6 & 44.0 \\
            FourierFT & 0.42 & 55.9 & 77.8 & 28.9 & 53.3 & 62.9 & 56.1 & 60.8 & 53.1 & 58.2 \\
            \cmidrule(lr){1-1} \cmidrule(lr){2-11}
            
            LoRA & 0.70 & 51.6 & 77.2 & 28.2 & 52.7 & 57.9 & 52.8 & 64.8 & 55.3 & 57.7 \\
            \textbf{SeLoRA$_{\text{F}}$} & 0.42 & 52.1 & 77.9 & 29.4 & 53.7 {\scriptsize (\textcolor{blue}{$\uparrow$ 1.0})} & \colorbox{backblue}{63.4} & \colorbox{backblue}{56.7} & 61.4 & 53.7 & 58.8 {\scriptsize (\textcolor{blue}{$\uparrow$ 1.1})} \\
            \textbf{SeLoRA$_{\text{W}}$} & 0.42 & 52.0 & \colorbox{backblue}{80.3} & \colorbox{backblue}{29.8} & \colorbox{backblue}{55.1 {\scriptsize (\textcolor{blue}{$\uparrow$ 2.4})}} & 59.3 & 55.6 & \colorbox{backblue}{\textbf{66.1}} & \colorbox{backblue}{\textbf{56.6}} & \colorbox{backblue}{59.4 {\scriptsize (\textcolor{blue}{$\uparrow$ 1.7})}} \\
            \cmidrule(lr){1-1} \cmidrule(lr){2-11}
            
            DoRA & 0.71 & 64.3 & 78.0 & 28.7 & 53.4 & 60.4 & 56.1 & 61.9 & 53.7 & 58.0 \\
            \textbf{SeDoRA$_{\text{F}}$} & 0.42 & 64.7 & 78.9 & 29.2 & 54.1 {\scriptsize (\textcolor{blue}{$\uparrow$ 0.7})} & 62.8 & \colorbox{backblue}{\textbf{57.3}} & 61.6 & 53.3 & 58.8 {\scriptsize (\textcolor{blue}{$\uparrow$ 0.8})} \\
            \textbf{SeDoRA$_{\text{W}}$} & 0.42 & 64.7 & \colorbox{backblue}{\textbf{80.4}} & \colorbox{backblue}{\textbf{30.3}} & \colorbox{backblue}{\textbf{55.4} {\scriptsize (\textcolor{blue}{$\uparrow$ \textbf{2.0}})}} & \colorbox{backblue}{\textbf{63.4}} & 56.1 & \colorbox{backblue}{63.5} & \colorbox{backblue}{55.3} & \colorbox{backblue}{\textbf{59.6 {\scriptsize \textcolor{blue}{(\textcolor{blue}{$\uparrow$ 1.6})}}}} \\
            
            \bottomrule
        \end{tabular}
    }
    \caption{Comparison of LLaMA2$_{\textsc{7b}}$ and LLaMA3$_{\textsc{8b}}$ against various methods on mathematical reasoning and code generation.
    We report the overall accuracy for mathematical reasoning and Pass@1 for code generation.
    The highest scores within each baseline method are highlighted in \colorbox{backblue}{blue} and the best results of each LLM are marked in \textbf{bold}.
    Memory is measured on 1x A100 GPU for mathematical reasoning per the micro-batch size in configurations.
    }
    \label{tab:math_code}
\end{table*}

\paragraph{Tasks.}
Our goal is to provide a rich picture of how our proposed approach performs in different scenarios.
Our experiments generally align with those reported by~\citet{liu2024dora} and~\citet{biderman2024lora}. 
We apply all the methods to instruction fine-tuning and evaluated their performance on conventional commonsense reasoning and two challenging tasks - mathematical reasoning and code generation. 
To provide a more rigorous evaluation, we adopt \texttt{alphaca-chat} prompt template throughout the training and assessments, more details are provided in Appendix~\ref{appendix:prompt}.

$\bullet$ \textbf{Commonsense Reasoning.} 
We utilize Commonsense170K~\cite{hu2023llm} as the training data, a collection of multiple-choice question-answer (QA) pairs derived from the training sets of eight sub-tasks: BoolQ~\citep{clark2019boolq}, PIQA~\citep{bisk2020piqa}, SIQA~\citep{sap2019socialiqa}, HellaSwag~\citep{zellers2019hellaswag}, WinoGrande~\citep{sakaguchi2021winogrande}, ARC-e, ARC-c~\citep{clark2018think}, and OBQA~\citep{mihaylov2018can}.
For evaluation, we employ greedy search for answer generation and assess the model's performance on the test sets of each dataset. 
Consistent with the protocols in~\citet{hu2023llm} and~\citet{liu2024dora}, the model's response is recorded as the first occurrence of the answer keywords in the generated output.

$\bullet$ \textbf{Mathematical Reasoning.} 
We employ MetaMathQA~\cite{yu2023metamath} for model fine-tuning, which consists of 395K mathematical QA pairs evolved from GSM8K~\cite{cobbe2021training} and MATH~\citep{hendrycks2020measuring}.
The evaluation is performed on the respective test sets of GSM8K and MATH, both of which require chain-of-thought reasoning~\citep{wei2022chain} to reach the final answer.
Following the evaluation protocol outlined in~\citet{yu2023metamath}, 
we assess performance by measuring the accuracy of the final numeric answer generated through greedy search.

$\bullet$ \textbf{Code Generation.} 
We utilize Magicoder-Evol-Instruct-110k~\cite{wei2024magicoder} as the training dataset, a programming QA collection that has been reproduced and decontaminated from WizardCoder~\cite{luo2024wizardcoder}. 
The fine-tuned models are evaluated on the HumanEval~\cite{chen2021evaluating} and MBPP~\cite{austin2021program} benchmarks. 
To ensure comprehensive evaluation, we also assess the models on HumanEval+ and MBPP+ using the evaluation protocol outlined in EvalPlus~\cite{liu2024your} and report the Pass@1 metric for each benchmark.

\paragraph{Baselines.}
We choose LoRA~\citep{hu2021lora}, DoRA~\citep{liu2024dora} and HiRA~\citep{huang2025hira} as the integrable baselines, while $\mathcal{LS}$-LoRA~\citep{he2022sparseadapter}, LoRETTA$_{rep}$~\cite{yang2024loretta} and FourierFT~\citep{gao2024parameter} are selected as PEFT methods with sparse re-parameterization.
All experiments are conducted using two open-source LFMs: LLaMA2$_{\textsc{7b}}$~\citep{touvron2023llama2} and LLaMA3$_{\textsc{8b}}$~\citep{dubey2024llama}.
Following common practice~\cite{kopiczko2023vera}, we used the base versions instead of the instruction-tuned ones.

\paragraph{Implementation Details.}
In line with the setup suggested in~\citet{hu2023llm}, we fix all baseline models to a rank of $r = 32$ and set $\alpha = 64$ while conducting hyperparameter search on learning rates employing AdamW optimizer~\cite{loshchilov2018decoupled} during fine-tuning.
To ensure fairness, we reproduce the results of LoRA, DoRA, FourierFT, $\mathcal{LS}$-LoRA and LoRETTA$_{rep}$ using their official implementations while implementing HiRA ourselves, as its official code is not publicly available, based on the optimal configurations reported in their original papers.
We apply PEFTs to the query, key, and value modules in attention ($\mathbf{W}_{\text{q}}, \mathbf{W}_{\text{k}}, \mathbf{W}_{\text{v}}$) and two feed-forward networks ($\mathbf{W}_{\text{up}}, \mathbf{W}_{\text{down}}$). 
For commonsense reasoning tasks, the LLMs are fine-tuned for 3 epochs, setting $\eta$ to 0.4 for LLaMA2$_{\textsc{7b}}$ and 0.6 for LLaMA3$_{\textsc{8b}}$.
For more complex tasks, we decrease $\eta$ to 0.2 for LLaMA2$_{\textsc{7b}}$ and 0.4 for LLaMA3$_{\textsc{8b}}$ with 2 fine-tuned epochs.
More details are provided in Appendix~\ref{appendix:exp}.

\begin{figure*}[t]
    \centering
    \subfigure[]{
        \includegraphics[width=.315\linewidth]{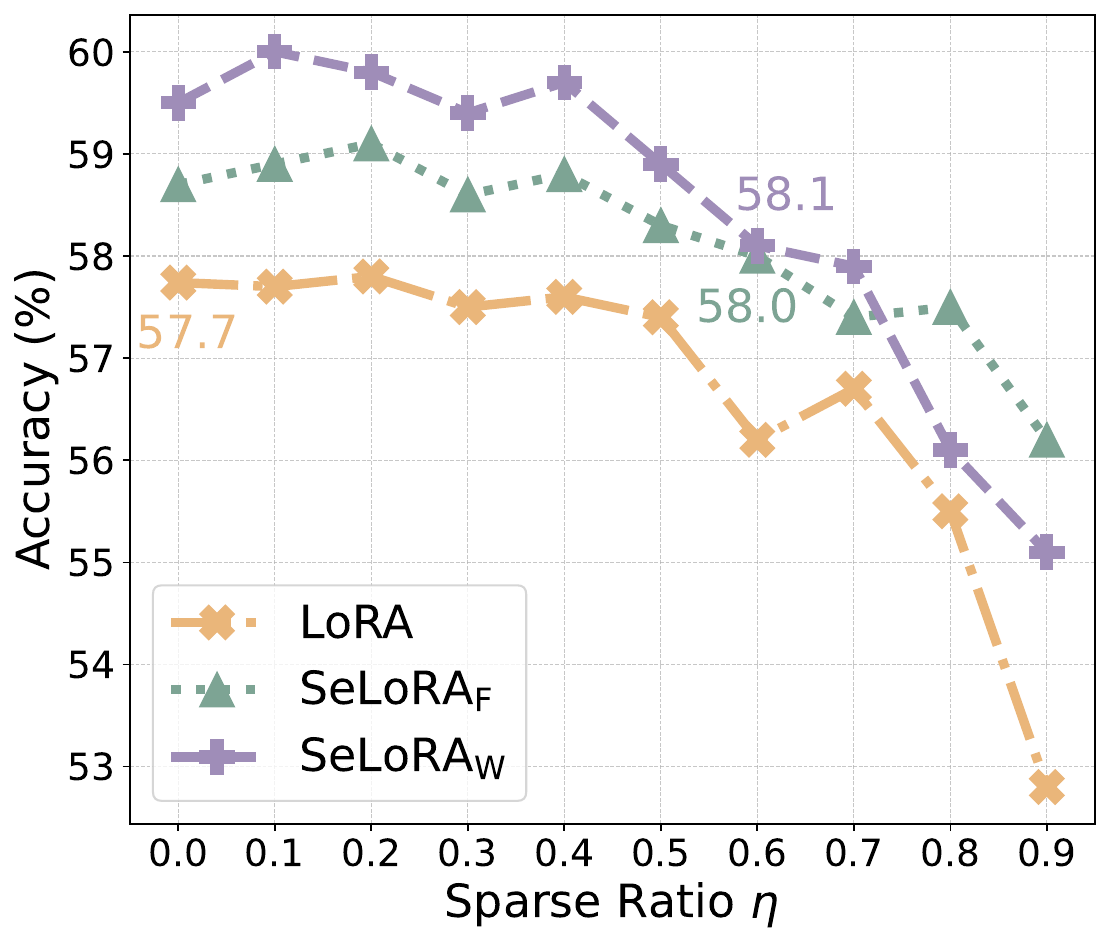}
        \label{fig:sparse_code}
    }
    \subfigure[]{
        \includegraphics[width=.315\linewidth]{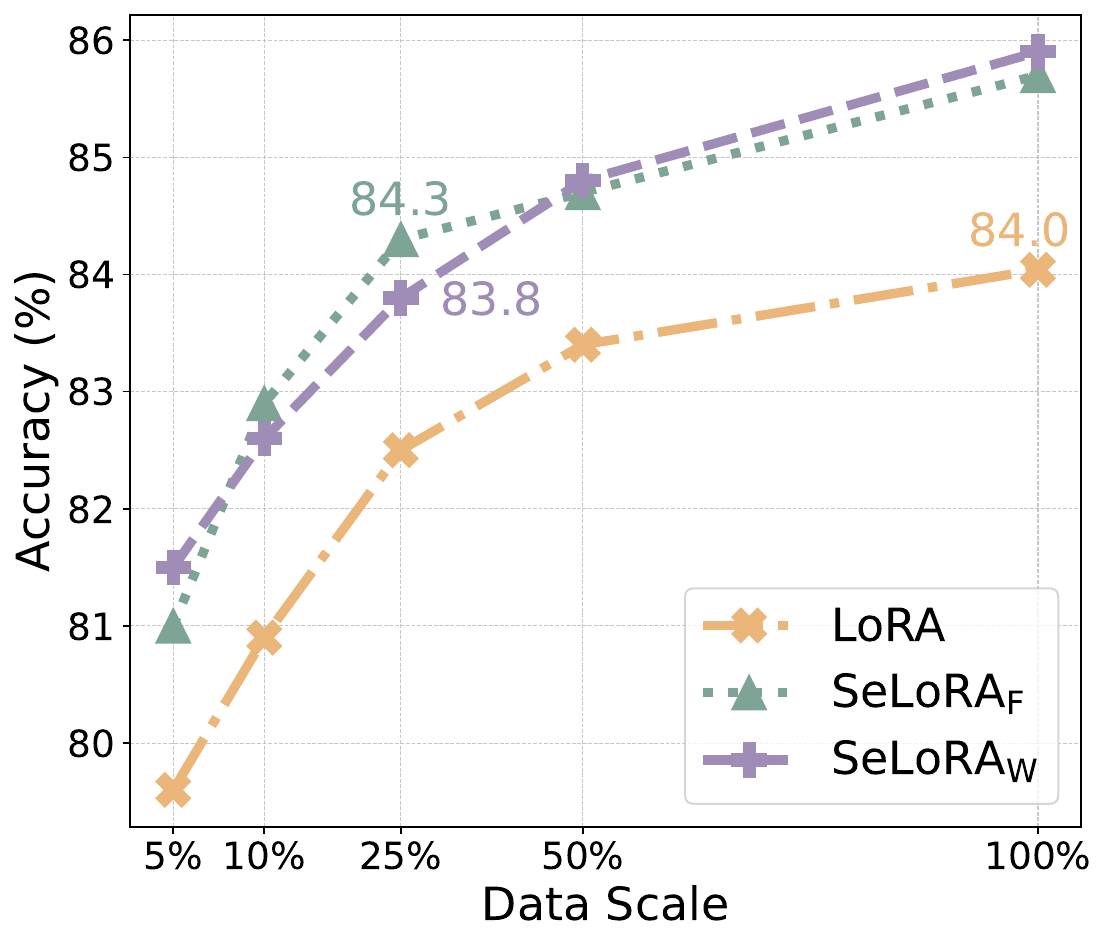}
        \label{fig:scale_dataset}
    }
    \subfigure[]{
        \includegraphics[width=.315\linewidth]{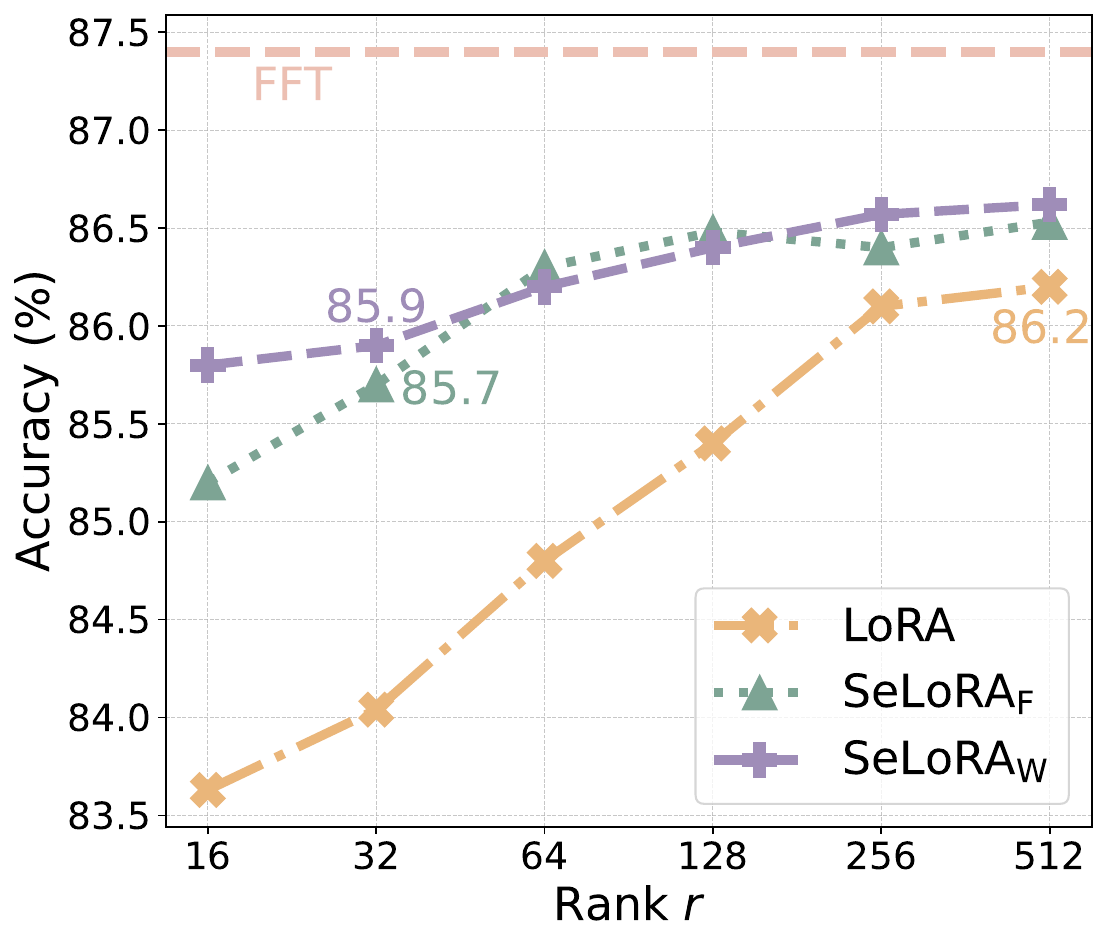}
        \label{fig:rank_increase}
    }
    \caption{
    (a) Average performance of code generation with varying sparse ratio $\eta$ on LLaMA3$_{\textsc{8b}}$; (b) Average performance of commonsense reasoning using different scales of training dataset; (c) Average performance of LLaMA3$_{\textsc{8b}}$ when the rank $r$ increases.
    }
    \label{fig:general_exp}
\end{figure*}

\subsection{Commonsense Reasoning}

Table~\ref{tab:common} provides a comprehensive overview of the general performance across different backbone architectures and baseline methods.
In comparison with $\mathcal{LS}$-LoRA and LoRETTA$_{rep}$, which employ sparse re-parameterizations, both variants of SeLoRA achieve better performance under a similar parameter budget while consuming less training time. 
As evident from the results, our proposed methods consistently surpass all integrable baselines in terms of average accuracy.
In particular, both SeLoRA$_{\text{F}}$ and SeLoRA$_{\text{W}}$ exhibit significant improvements over LoRA, achieving an average accuracy gain of approximately +2.0 across the LLaMA families.
While integrating our methods into more advanced baselines such as DoRA and HiRA results in slightly reduced gains, our methods still deliver notable improvements, reaching up to +2.1 on LLaMA2${_\textsc{7b}}$ and +1.3 on LLaMA3${_\textsc{8b}}$.
These results underscore the adaptability of our approach as a plug-and-play framework. 
Moreover, our methods enhance learning capacity while requiring significantly fewer trainable parameters, all without increasing training time, underscoring their efficiency.
Additionally, we observe that parameterizing the weight matrix using wavelet transformations generally leads to more pronounced performance gains.
This can be attributed to the wavelet-based representation in better balancing smoothness with detail preservation.

\subsection{More Challenging Tasks}

Table~\ref{tab:math_code} presents the overall results for mathematical reasoning and code generation.
The zero-shot performances for code generation are in line with EvalPlus leaderboard\footnote{https://evalplus.github.io/leaderboard.html}.
Consistent with our findings in commonsense reasoning, our approaches consistently surpass LoRA and DoRA among LLaMA families for more challenging tasks.
In comparison with FourierFT, which also leverages Fourier transformations, SeLoRA${\text{F}}$ achieves comparable performance under a similar parameter budget while exhibiting significantly lower memory consumption. 
This advantage arises because $\Delta \mathbf{W}$ of FourierFT has the same shape as the pre-trained weights, whereas SeLoRA$_{\text{F}}$ retains the low-rank update structure. 
Moreover, wavelet encoding continues to provide stable and substantial improvements, whereas fourier encoding exhibits greater performance fluctuations with a reduced level of enhancement.
Particularly, wavelet-based variants achieve average improvements of +2.1 and +1.9 across all configurations for mathematical reasoning and code generation respectively.
These results, in conjunction with our observations in commonsense reasoning, substantiate the effectiveness of our proposed method across diverse reasoning and generation tasks.
\section{In-depth Analyses}\label{sec:analysis}

In this section, we conduct a variety of quantitive analyses on our proposed approach to assess its robustness and generalizability.
The experimental implementations adhere to the setup in Section~\ref{sec:setup} unless otherwise specified.

\paragraph{Sparsity Utilization.}
To further investigate the impact of spectral encoding in leveraging the \textit{sparsity property}, we evaluate SeLoRA under two scenarios: \textbf{(1)} varying the sparse ratio $\eta$ from 0.0 to 0.9 and \textbf{(2)} adjusting the rank $r$ while keeping parameter counts fixed.

As presented in Figure~\ref{fig:motivation} and Figure~\ref{fig:sparse_code}, the maximum sparse ratio that maintains comparable expressive power varies across tasks and backbone models.
More challenging tasks and less expressive backbones have smaller values and exhibit lower redundancy.
Nevertheless, both SeLoRA variants consistently outperform LoRA across all tasks and backbone architectures.
Notably, SeLoRA maintains performance parity with LoRA even at relatively high sparse ratios, such as $\eta = 0.8$ for commonsense reasoning and $0.6$ for code generation in LLaMA3$_{\textsc{8b}}$.
Moreover, the performance gains are already pronounced at high sparse ratios, which can be attributed to the intrinsic properties of spectral transformations that allow high-quality reconstruction with extremely sparse elements.

\begin{figure}[t]
    \includegraphics[width=\linewidth]{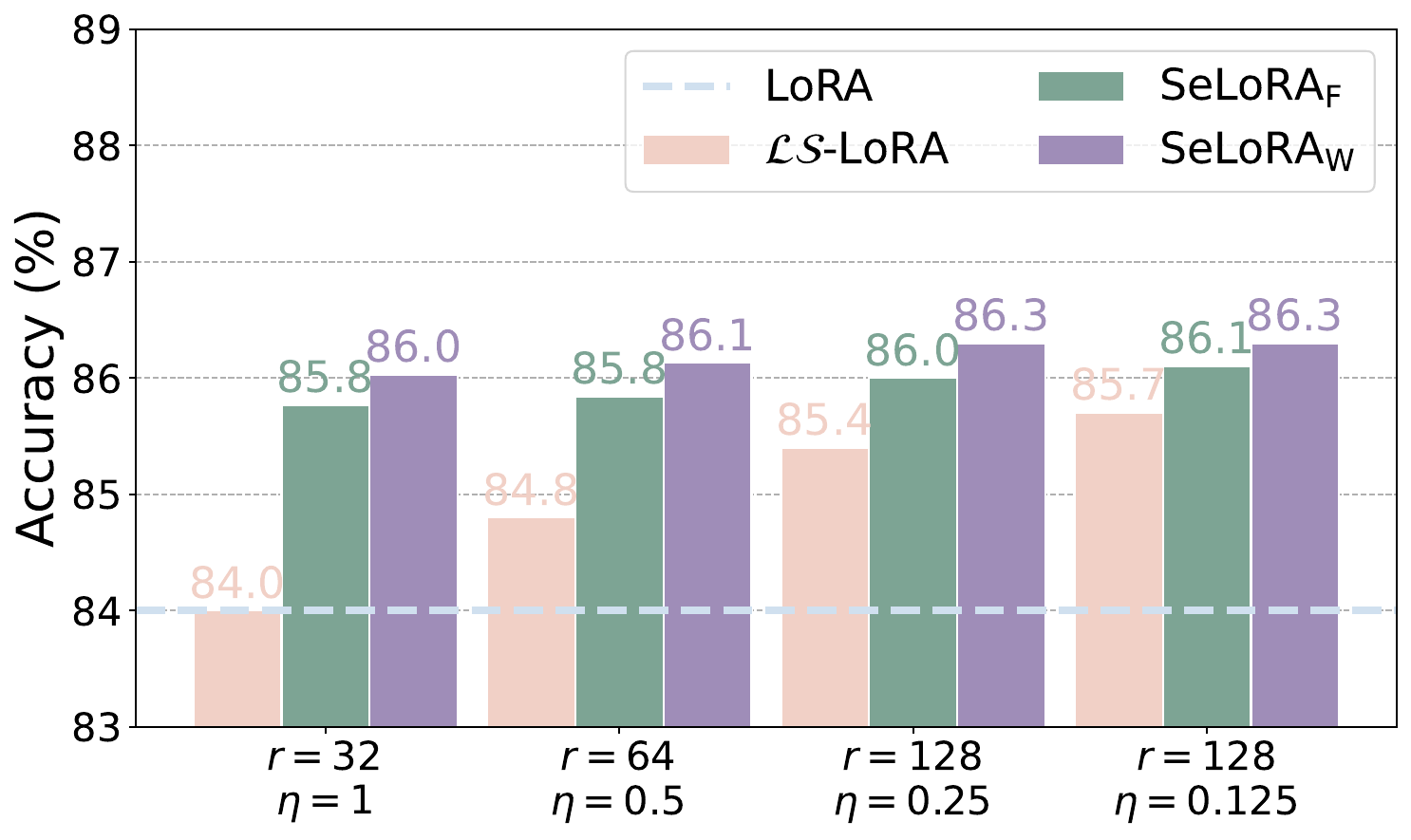}
    \caption{
    Performance distribution with different sparse learning mechanisms on LoRA modules for LLaMA3$_\textsc{8b}$ on commonsense reasoning.
    $\mathcal{LS}$-LoRA achieves a higher rank $r$ while maintaining the same parameter budget against trivial LoRA via a proper sparse ratio $\eta$.
    }
    \label{fig:sparsification}
\end{figure}

Furthermore, as illustrated in Figure~\ref{fig:sparsification}, pruning-based $\mathcal{LS}$-LoRA enhances performance by utilizing higher ranks through masked adaptation under fixed parameter budgets. SeLoRA further amplifies these gains by combining spectral encoding with masking, consistently delivering superior results.
These findings jointly substantiate the significance of employing spectral encoding in SeLoRA for effectively harnessing the \textit{sparsity property}.

\paragraph{Data Scalability.}
We explore the influence of training data size on the performance of our approach.
We experiment with the LLaMA3$_{\textsc{8b}}$ employing the rank of $r = 32$.
For comprehensiveness, the examination involves random sampling of 5\%, 10\%, 25\%, and 50\% instances from the training data for the commonsense reasoning task.
As illustrated in Figure~\ref{fig:scale_dataset} and Table~\ref{tab:common_train_ratio}, both variants of our approach exhibit steadily increased performance with an increase in training data volume.
Impressively, with just 25\% of the training data, SeLoRA outperforms LoRA even when the latter utilizes the entire dataset, highlighting SeLoRA's exceptional efficiency in leveraging training data for performance improvement.

\begin{table}[t]
    \centering
    \resizebox{.95\linewidth}{!}{
        \begin{tabular}{l*{5}{c}}
            \toprule
            & & \multicolumn{3}{c}{Task} \\
            \cmidrule(lr){3-5}
            Methods & Bases & Common & Math & Code & Avg. \\
            \midrule 
            LoRA & - & 83.9 & 52.7 & 57.6 & 64.7 \\
            \midrule[0.6pt]
            \midrule[0.6pt]
            \multirow{4}{*}{SeLoRA$_{\text{W}}$}
            & Haar & 85.9 & 55.1 & 59.4 & 66.8  \\
            & Db & 85.9 & \textbf{55.4} & 59.1 & 66.8 \\
            & Bior & 85.9 & 54.8 & 59.5 & 66.7 \\
            & Coif & \textbf{86.2} & 55.2 & \textbf{59.8} & \textbf{67.0} \\
            \bottomrule
        \end{tabular}
    }
    \caption{Peformance variations with different wavelet filters on LLaMA3$_{\textsc{8b}}$.
    }
    \label{tab:wavelet_ablation}
\end{table}

\paragraph{Rank Scalability.}
We assess the sensitivity of our proposed approach against different ranks $r$ with an increasing rank sequence [16, 32, 64, 128, 256, 512], as higher ranks are known to improve performance on complex tasks~\citet{biderman2024lora}.
In alignment with Section~\ref{sec:setup}, we conduct evaluations on LLaMA3$_{\textsc{8b}}$ for commonsense reasoning while keeping $\eta = 0.6$ for SeLoRA.
As illustrated in Figure~\ref{fig:rank_increase} and Table~\ref{tab:common_rank}, SeLoRA exhibits a similar trend to LoRA, benefiting from an expanded learning space and achieving progressively better performance as $r$ increases while maintaining sparsity.
Notably, we observe diminishing performance gains for LoRA beyond $r=256$, whereas SeLoRA reaches comparable performance at just $r=32$.
This indicates that spectral encoding effectively enhances LoRA’s efficiency, enabling it to realize the benefits of higher ranks with significantly fewer trainable parameters.

\begin{figure}[t]
    \includegraphics[width=\linewidth]{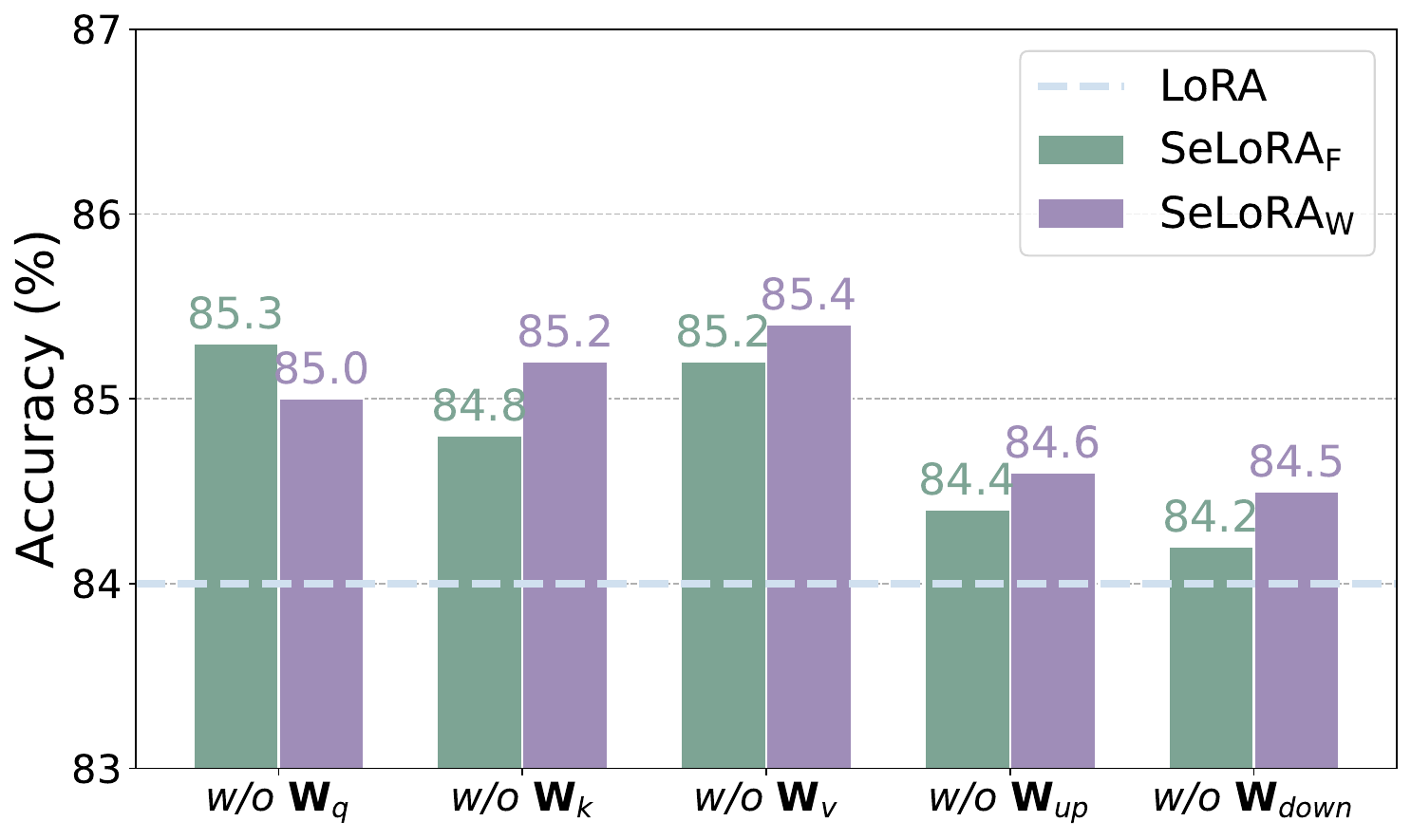}
    \caption{
    Performance distribution with the removal of spectral encoding among different modules for LLaMA3$_\textsc{8b}$ on commonsense reasoning. 
    }
    \label{fig:module}
\end{figure}

\paragraph{Module Sensitivity.}
Understanding which modules benefit most from spectral encoding is crucial for demystifying SeLoRA’s effectiveness.
Thereafter, we assess its impact by selectively removing spectral encoding from individual modules within $\{ \mathbf{W}_{\text{q}}, \mathbf{W}_{\text{k}}, \mathbf{W}_{\text{v}}, \mathbf{W}_{\text{up}}, \mathbf{W}_{\text{down}} \}$ yet keeping all other modules unchanged.
As presented in Figure~\ref{fig:module}, the removal of spectral encoding from the feed-forward networks  ($\mathbf{W}_{\text{up}}, \mathbf{W}_{\text{down}}$) leads to a substantial performance drop, whereas its absence in the attention components ($\mathbf{W}_{\text{q}}, \mathbf{W}_{\text{k}}, \mathbf{W}_{\text{v}}$) results in relatively less degradation.
This indicates parameters of feed-forward networks benefit more significantly from strategies of enhancing parameter utilization, which also aligns with insights from previous studies~\citep{biderman2024lora, jiang2024taia}.

\paragraph{Basis Expressiveness.}
We examine the effect of different wavelet bases on our approach, as they offer varying trade-offs between smoothness and detail preservation.
In addition to the Haar wavelet, we explore three alternative bases: Daubechies-4 (Db), Biorthogonal (Bior), and Coiflets (Coif).
As reported in Table~\ref{tab:wavelet_ablation}, SeLoRA consistently outperforms LoRA across various tasks, regardless of the specific wavelet transformation used, with only minor performance variations among the different wavelet bases.
This demonstrates SeLoRA’s robustness to wavelet selection.

\begin{figure}[t]
    \centering
    \includegraphics[width=\linewidth]{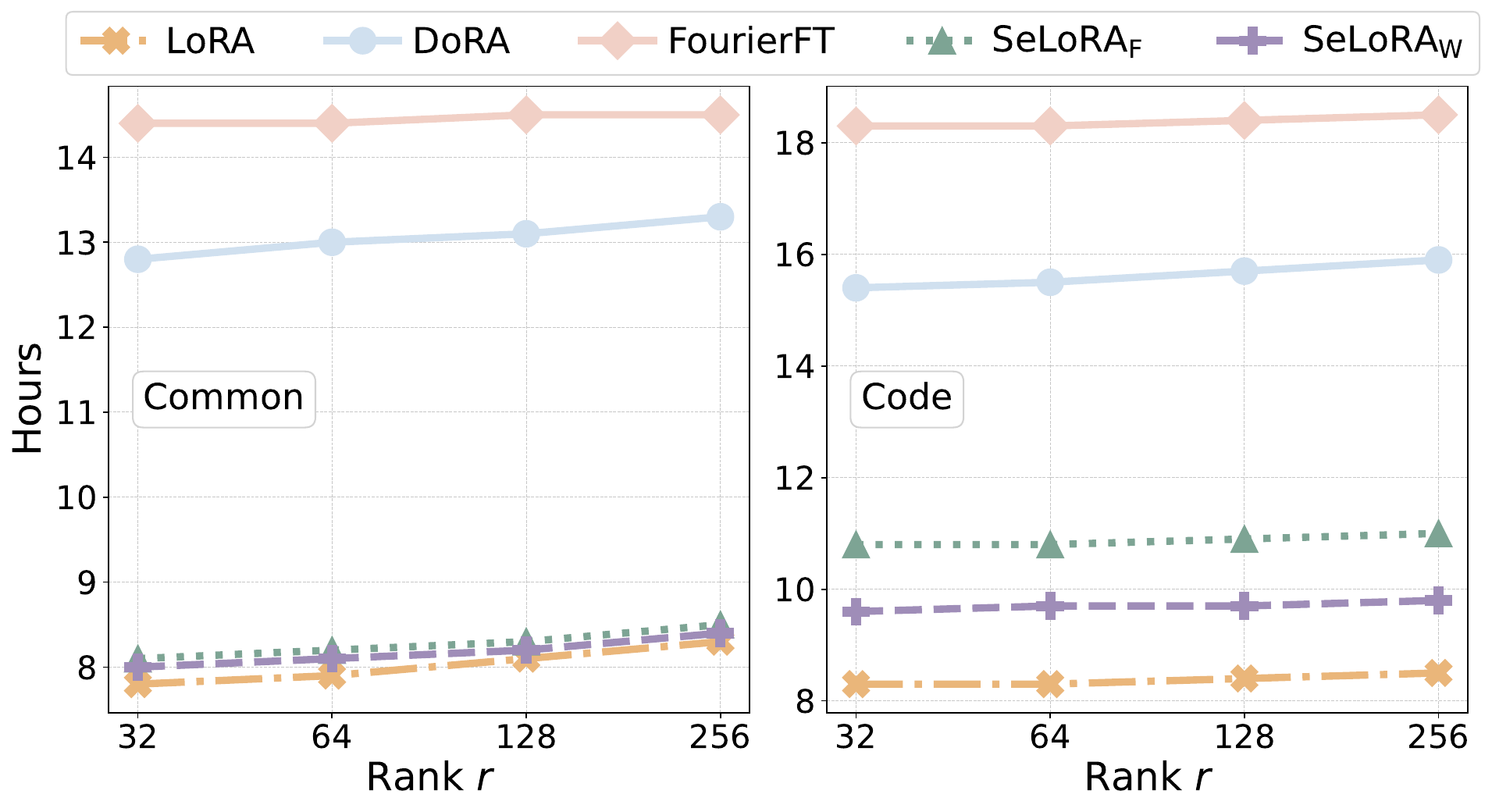}
    \caption{Comparison of training times across different methods adapted to LLaMA3$_{\textsc{8b}}$.}
    \label{fig:efficiency}
\end{figure}

\paragraph{Efficiency Comparison.}
We evaluate the training efficiency of our approach against different methods when adapting at varying ranks $r$.
For consistency, all evaluations are conducted on 1x A100 GPU and FourierFT is assigned a comparable parameter budget to SeLoRA at each rank.
As illustrated by Figure~\ref{fig:efficiency} and Table~\ref{tab:common} and~\ref{tab:math_code}, SeLoRA achieves competitive performance compared to DoRA and FourierFT while requiring substantially less training time, and remains comparable in efficiency to LoRA.
This highlights SeLoRA’s advantage in balancing performance and computational cost.

\paragraph{Subspace Analysis.}
Inspired by~\citet{hu2021lora}, we investigate the correlation between $\mathbf{W}$ and $\Delta \mathbf{W}$.
We answer this question by projecting $\mathbf{W}$ onto the r-dimensional subspace of $\Delta \mathbf{W}$ by computing $\mathbf{U}^{T}_{r} \mathbf{W} \mathbf{V}_{r}$, with $\mathbf{U}/\mathbf{V}$ being the left/right singular-vector matrices of $\Delta \mathbf{W}$. 
As defined in~\citet{jiang2025finetuning}, the amplification factor (AF) $\frac{||\Delta \mathbf{W}||_{F}}{||\mathbf{U}^{T}_{r} \mathbf{W} \mathbf{V}_{r}||_{F}}$  measures the subspaces emphasized in the $\Delta \mathbf{W}$ when compared with $\mathbf{W}$, while the reverse amplification factor (RAF) $\frac{||\Delta \mathbf{W}||_{F}}{||\mathbf{U}^{T}_{d - r} \mathbf{W} \mathbf{V}_{d - r}||_{F}}$ indicates the already-amplified directions of $\mathbf{W}$ not being activated.
From Table~\ref{tab:subspace}, we can draw the following conclusions: 
1) Both LoRA and SeLoRA amplify the important features learned from the tasks but not emphasized in $\mathbf{W}$;
2) Compared to LoRA, SeLoRA further reduces the amplification of already-emphasized features in $\Delta \mathbf{W}$.

\begin{table}[t]
    \centering
    \resizebox{\linewidth}{!}{
        \begin{tabular}{l*{4}{c}}
            \toprule
            Methods & $||\Delta \mathbf{W}||_{F}$ & $||\mathbf{U}^{T}_{r} \mathbf{W} \mathbf{V}_{r}||_{F}$ & AF ($\uparrow$) & RAF ($\downarrow$) \\
            \midrule 
            LoRA & 2.74 & 1.29 & 2.13 & 0.17 \\
            \midrule[0.6pt]
            \midrule[0.6pt]
            SeLoRA$_{\text{F}}$ & 1.98 & 0.99 & 2.01 & 0.09 \\
            SeLoRA$_{\text{W}}$ & 2.11 & 0.89 & \textbf{2.37} & \textbf{0.04} \\
            \bottomrule
        \end{tabular}
    }
    \caption{
     The amplification factor and reverse amplification factor. The weight matrices are taken from the 24$^{th}$ layer of LLaMA3$_\textsc{8b}$ trained for code generation.
    }
    \label{tab:subspace}
\end{table}
\section{Conclusion}\label{sec:conclusion}

In this paper, we first explore the impact of parameter redundancy in LoRA fine-tuning, revealing sparse tunable parameters are sufficient for expressive learning, a phenomenon termed as \textit{sparsity property}.
Built on these insights, we present SeLoRA, a novel extension of LoRA that leverages spectral transformations to re-parameterize adaptation matrices, enhancing parameter efficiency without compromising training or inference performance.
With its flexible and modular nature, SeLoRA can be seamlessly integrated into various LoRA variants as a plug-and-play framework.
Empirical evaluations demonstrate that SeLoRA delivers superior adaptability and improvements across diverse instruction tuning tasks. 
Further in-depth investigations validate its robustness and provide insights into its underlying mechanisms, highlighting its feasibility and practical utility.
Future work may explore more effective strategies for leveraging parameter redundancy, further unlocking LoRA’s potential for enhanced efficiency and scalability.
\section*{Limitations}

We observe that the performance improvements of SeLoRA over LoRA gradually diminish as the rank increases, ultimately converging to a similar upper bound at higher ranks. 
This suggests that while SeLoRA effectively leverages spectral transformation to achieve the expressiveness of high-rank LoRA in a more parameter-efficient manner, it remains constrained by the inherent capacity limits of LoRA itself.
Furthermore, SeLoRA is currently compatible with LoRA variants that follow the same update schema. 
However, its integration into alternative update strategies, such as SVD-based decomposition, remains underexplored.
Lastly, due to computational constraints, we are unable to evaluate SeLoRA on models with 70 billion parameters. 
Exploring its scalability on larger models is an important direction for future work.
\section*{Acknowledgements}

This work was supported by National Natural Science Foundation of China Grant No. 72371217, the Guangzhou Industrial Informatic and Intelligence Key Laboratory No. 2024A03J0628, the Nansha Key Area Science and Technology Project No. 2023ZD003, and Project No. 2021JC02X191.

\bibliography{anthology,acl2025}

\clearpage
\appendix



\section{Related Works}\label{sec:related}

\subsection{Parameter-Efficient Fine-Tuning}

Fine-tuning large pre-trained language models is crucial for improving NLP tasks~\citep{chen2023large, chen2024controlmath, chen2024graphwiz}. 
However, updating all model parameters is computationally intensive and storage-demanding for models like GPT-3~\citep{brown2020language} and LLaMA~\citep{touvron2023llama}. 
Parameter-efficient fine-tuning (PEFT) methods address these issues by updating fewer parameters or adding lightweight modules.

One prominent approach in PEFT is the use of adapters, namely, small bottleneck layers inserted within each layer of a pre-trained model~\citep{houlsby2019parameter, pfeiffer2020adapterfusion, karimi2021compacter, he2021towards}. 
\citet{houlsby2019parameter} introduced adapters that enable task-specific adaptation while keeping the original model weights fixed. 
Building upon this, \citet{pfeiffer2020adapterfusion} proposed a modular adapter framework that facilitates multi-task transfer.
To further optimize parameter efficiency, \citet{karimi2021compacter} reduced the number of parameters by employing parameter sharing and low-rank approximations within adapters. 
Another line of research involves prompt tuning, which modifies the input embeddings to guide the model toward specific tasks~\citep{lester2021power, liu2021p, li2021prefix, chen2023understanding}. 
\citet{lester2021power} optimized continuous prompt embeddings while keeping the language model's parameters fixed, demonstrating the effectiveness of prompt tuning for task adaptation. 
Similarly, Prefix-Tuning~\citep{li2021prefix} prepends trainable vectors to the input of each transformer layer without altering the model architecture, effectively steering the model toward desired behaviors with minimal parameter updates.

While these methods exhibit high efficiency and preserve the originality of the pre-trained model, they inevitably introduce higher inference costs due to additional modules or modifications required during deployment. 
In contrast, LoRA~\citep{hu2021lora} and its variants~\citep{zhang2023lora, balazy2024lora, li2024vb, liu2024dora, nikdan2024rosa, gao2024parameter, jiang2024mora, huang2025hira} inject trainable low-rank matrix decomposition into transformer layers, have been widely used to adapt recent LFMs for various tasks. 
These approaches not only reduce the number of trainable parameters but also enable seamless merging with the original model weights, thereby avoiding increased inference burdens.
Among these studies,~\citet{zhang2023adalora} and~\citep{benedek2024prilora} further highlight that LoRA's intrinsic properties can be further exploited to enhance learning efficiency.
In this work, we revisit the inherent parameter redundancy in LoRA and leverage this property to further unlock its potential.

\subsection{Parameter Redundancy}

Parameter redundancy has been extensively observed in pre-trained language models~\citep{dalvi2020analyzing, bhojanapalli2021leveraging, he2022sparseadapter}, and recent studies have demonstrated that this redundancy can be exploited to accelerate inference speed~\citep{he2024matters, men2024shortgpt}.
Parallel findings reveal that parameter redundancy also exists within the context of LoRA for LFMs, presenting opportunities to enhance efficiency from two complementary perspectives:
\textbf{(1)} Pre-processing: This research direction focuses on reducing the number of learnable parameters during the fine-tuning phase without sacrificing model performance.
Various techniques have been proposed, including matrix projection~\citep{kopiczko2023vera, renduchintala2023tied}, matrix decomposition~\citep{li2024vblora}, and matrix substitution~\citep{sehanobish2024structured, chen2024parameter}.
\textbf{(2)} Post-processing: This line of work aims to remove redundant parameters after fine-tuning to improve the model's capacity and adaptability.
Notable applications include efficient model merging~\citep{yu2024language}, out-of-domain adaptation~\citep{jiang2024taia}, and enhancing learning capacity~\citep{jiang2025finetuning}.

Despite these advances, pre-processing approaches for parameter reduction often rely on heuristic designs, lacking a systematic understanding of the specific impact of parameter redundancy during the fine-tuning phase of LoRA for LFMs.
In this work, we aim to bridge this gap by providing a comprehensive investigation of parameter redundancy in LoRA fine-tuning and leveraging it to achieve more efficient and expressive fine-tuning strategies.

\subsection{Sparse Learning}

Sparse neural networks exploit the fact that many weights in over-parameterized models can be pruned with minimal impact on performance~\citep{han2015learning, lee2018snip, frankle2018lottery, wang2020picking, liu2022unreasonable, frantar2023sparsegpt}. 
A common technique, magnitude pruning~\citep{han2015deep}, eliminates weights with small magnitudes, significantly reducing model size while maintaining performance. 
Meanwhile, dynamic sparsity methods~\citep{mocanu2018scalable, zhang2022advancing, chen2023deepzero} iteratively adjust the sparsity patterns during training, allowing the network to discover efficient architectures on the fly adaptively. 

Another promising direction involves learning in transformed domains, such as spectral space.
By leveraging expressive spectral transformations~\citep{cheng2023wiener}, parameterizing weight matrices with sparse, learnable components in the spectral domain has been shown to retain strong expressiveness in both traditional neuroevolution~\citep{koutnik2010evolving, van2016wavelet} and modern neural networks~\citep{wolter2020neural, irie2021training}.
In light of the inherent characteristics of LoRA, this work aims to investigate the potential of this simple yet powerful design to further enhance its efficiency.

\section{Wavelet Construction}\label{appedix:wavelet}

Different wavelet bases strike a balance between smoothness and detail preservation. 
To illustrate the construction process, we use the Haar wavelet as an example. 
The 2D Haar wavelet filter is derived by extending the 1D Haar wavelet through tensor products.

In the 1D case, the Haar wavelet is defined by a scaling function, $\phi(\cdot)$, and a wavelet function, $\psi(\cdot)$, given by:
\begin{equation}
    \phi(x) = 
    \begin{cases}
        1, & 0 \le x <1, \\
        0, & \text{otherwise}, 
    \end{cases}
\end{equation}
with
\begin{equation}
    \psi(x) = 
    \begin{cases}
        1, & 0 \le x < 0.5, \\
        -1 & 0.5 \le x < 1, \\
        0, & \text{otherwise}.
    \end{cases}
\end{equation}

The 2D Haar wavelet basis functions are then constructed via tensor products as follows:
\begin{equation}
\begin{aligned}
    \psi^{a}(x, y) = \phi(x)\phi(y), \\
    \psi^{h}(x, y) = \psi(x)\phi(y), \\
    \psi^{v}(x, y) = \phi(x)\psi(y), \\
    \psi^{d}(x, y) = \psi(x)\psi(y),
\end{aligned}
\end{equation}
where the superscripts denote different coefficient types as defined in Section~\ref{sec:method}. 
This construction method can be generalized to other wavelets by applying the same procedure using their respective 1D scaling and wavelet functions.

\section{Additional Experimental Details}\label{appendix:exp}

\paragraph{Training Configurations.}
All our experiments were carried out on Linux servers equipped with an AMD EPYC 7763 64-core CPU processor, 512GB RAM, and 8x NVIDIA A100 80G GPU with BFloat16 precision.
The detailed configurations for all instruction-tuning tasks are illustrated in Table~\ref{tab:common_hyperparameters}, Table~\ref{tab:meta_hyperparameters}, and Table~\ref{tab:code_hyperparameters}.

\paragraph{Additional Results.}
The full experimental results in Section~\ref{sec:analysis} are listed in Table~\ref{tab:common_train_ratio} and Table~\ref{tab:common_rank}.

\section{Prompt Template}\label{appendix:prompt}

\begin{prompt}{Training Prompt}{trainprompt}
{
\small
Below is an instruction that describes a task. 
Write a response that appropriately completes the request. \\

\#\#\# Instruction: \\
\{Question\} \\

\#\#\# Response: \\
\{Answer\}

}
\end{prompt}

\begin{prompt}{Evaluation Prompt (Common)}{evalprompt_common}
{
\small
Below is an instruction that describes a task.
Write a response that appropriately completes the request. \\

\#\#\# Instruction: \\
\{Question\} \\

\#\#\# Response: 

}
\end{prompt}

\begin{prompt}{Evaluation Prompt (Math)}{evalprompt_math}
{
\small
Below is an instruction that describes a task.
Write a response that appropriately completes the request. \\

\#\#\# Instruction: \\
\{Question\} \\

\#\#\# Response: \\
Let's think step by step.

}
\end{prompt}

\begin{prompt}{Evaluation Prompt (Code)}{evalprompt_code}
{
\small
Below is an instruction that describes a task.
Write a response that appropriately completes the request. \\

\#\#\# Instruction: \\
\{Question\} \\

\#\#\# Response: \\
\{Import Section\} \\

\{Function Signature\} \\
\{Docstring\}

}
\end{prompt}

\begin{table*}[ht]
    \centering
    \resizebox{\textwidth}{!}{
        \begin{tabular}{l|*{12}{c}}
            \toprule
            & \multicolumn{6}{c}{LLaMA2$_{\textsc{7b}}$} & \multicolumn{6}{c}{LLaMA3$_{\textsc{8b}}$} \\
            \cmidrule(lr){2-7} \cmidrule(lr){8-13} 
            Hyperparameter & SeLoRA$_{\text{F}}$ & SeLoRA$_{\text{W}}$ & SeDoRA$_{\text{F}}$ & SeDoRA$_{\text{W}}$ & SeHiRA$_{\text{F}}$ & SeHiRA$_{\text{W}}$ & SeLoRA$_{\text{F}}$ & SeLoRA$_{\text{W}}$ & SeDoRA$_{\text{F}}$ & SeDoRA$_{\text{W}}$ & SeHiRA$_{\text{F}}$ & SeHiRA$_{\text{W}}$ \\
            \midrule
            Optimizer & \multicolumn{12}{c}{AdamW} \\
            LR Scheduler & \multicolumn{12}{c}{Cosine} \\
            Batch Size & \multicolumn{12}{c}{16} \\
            Warmup Steps & \multicolumn{12}{c}{100} \\
            Dropout & \multicolumn{12}{c}{0.05} \\
            Epochs & \multicolumn{12}{c}{3} \\
            Rank $r$ & \multicolumn{12}{c}{32} \\
            Alpha $\alpha$ & \multicolumn{12}{c}{64} \\
            Sparse Ratio $\eta$ & \multicolumn{6}{c}{0.4} & \multicolumn{6}{c}{0.6} \\
            Modules & \multicolumn{12}{c}{[q\_proj, k\_proj, v\_proj, up\_proj, down\_proj]} \\
            Learning Rate & 3e-4 & 2e-4 & 2e-4 & 2e-4 & 1e-3 & 1.5e-3 & 2e-4 & 2e-4 & 2e-4 & 2e-4 & 1.5e-3 & 1.5e-3 \\
            \bottomrule
        \end{tabular}
    }
    \caption{Hyperparameter configurations for commonsense reasoning.}
    \label{tab:common_hyperparameters}
\end{table*}
\begin{table*}[ht]
    \centering
    \resizebox{.9\linewidth}{!}{
        \begin{tabular}{l|*{8}{c}}
            \toprule
            & \multicolumn{4}{c}{LLaMA2$_{\textsc{7b}}$} & \multicolumn{4}{c}{LLaMA3$_{\textsc{8b}}$} \\
            \cmidrule(lr){2-5} \cmidrule(lr){6-9} 
            Hyperparameter & SeLoRA$_{\text{F}}$ & SeLoRA$_{\text{W}}$ & SeDoRA$_{\text{F}}$ & SeDoRA$_{\text{W}}$ & SeLoRA$_{\text{F}}$ & SeLoRA$_{\text{W}}$ & SeDoRA$_{\text{F}}$ & SeDoRA$_{\text{W}}$ \\
            \midrule
            Optimizer & \multicolumn{8}{c}{AdamW} \\
            LR Scheduler & \multicolumn{8}{c}{Cosine} \\
            Micro Batch Size & \multicolumn{8}{c}{32} \\
            Batch Size & \multicolumn{8}{c}{128} \\
            Warmup Ratio & \multicolumn{8}{c}{0.03} \\
            Dropout & \multicolumn{8}{c}{0.05} \\
            Epochs & \multicolumn{8}{c}{2} \\
            Rank $r$ & \multicolumn{8}{c}{32} \\
            Alpha $\alpha$ & \multicolumn{8}{c}{64} \\
            Sparse Ratio $\eta$ & \multicolumn{4}{c}{0.2} & \multicolumn{4}{c}{0.4} \\
            Modules & \multicolumn{8}{c}{[q\_proj, k\_proj, v\_proj, up\_proj, down\_proj]} \\
            Learning Rate & \multicolumn{4}{c}{3e-4} & \multicolumn{4}{c}{4e-4} \\
            \bottomrule
        \end{tabular}
    }
    \caption{Hyperparameter configurations for mathematical reasoning.}
    \label{tab:meta_hyperparameters}
\end{table*}
\begin{table*}[ht]
    \centering
    \resizebox{.9\textwidth}{!}{
        \begin{tabular}{l|*{8}{c}}
            \toprule
            & \multicolumn{4}{c}{LLaMA2$_{\textsc{7b}}$} & \multicolumn{4}{c}{LLaMA3$_{\textsc{8b}}$} \\
            \cmidrule(lr){2-5} \cmidrule(lr){6-9} 
            Hyperparameter & SeLoRA$_{\text{F}}$ & SeLoRA$_{\text{W}}$ & SeDoRA$_{\text{F}}$ & SeDoRA$_{\text{W}}$ & SeLoRA$_{\text{F}}$ & SeLoRA$_{\text{W}}$ & SeDoRA$_{\text{F}}$ & SeDoRA$_{\text{W}}$ \\
            \midrule
            Optimizer & \multicolumn{8}{c}{AdamW} \\
            LR Scheduler & \multicolumn{8}{c}{Cosine} \\
            Micro Batch Size & \multicolumn{8}{c}{8} \\
            Batch Size & \multicolumn{8}{c}{128} \\
            Warmup Ratio & \multicolumn{8}{c}{0.03} \\
            Dropout & \multicolumn{8}{c}{0.05} \\
            Epochs & \multicolumn{8}{c}{2} \\
            Rank $r$ & \multicolumn{8}{c}{32} \\
            Alpha $\alpha$ & \multicolumn{8}{c}{64} \\
            Sparse Ratio $\eta$ & \multicolumn{4}{c}{0.2}& \multicolumn{4}{c}{0.4} \\
            Modules & \multicolumn{8}{c}{[q\_proj, k\_proj, v\_proj, up\_proj, down\_proj]} \\
            Learning Rate & 4e-4 & 4e-4 & 4e-4 & 6e-4 & 4e-4 & 4e-4 & 6e-4 & 4e-4 \\
            \bottomrule
        \end{tabular}
    }
    \caption{Hyperparameter configurations for code generation.}
    \label{tab:code_hyperparameters}
\end{table*}

\section{Generation Examples}\label{appendix:example}

\begin{exmp}{Hellas.}{commonexample}
{
\small
Below is an instruction that describes a task.
Write a response that appropriately completes the request. \\

\#\#\# Instruction: \\
\textbf{\textcolor{lightblue}{
Please choose the correct ending to complete the given sentence: Roof shingle removal: A man is sitting on a roof. he \\
\\
Ending1: is using wrap to wrap a pair of skis. Ending2: is ripping level tiles off. Ending3: is holding a rubik's cube. Ending4: starts pulling up roofing on a roof. \\
\\
Answer format: ending1/ending2/ending3/ending4}} \\

\#\#\# Response: \\
\textbf{\textcolor{lightblue}{
the correct answer is ending4.
}}

}
\end{exmp}

\begin{exmp}{MATH}{mathexample}
{
\small
Below is an instruction that describes a task.
Write a response that appropriately completes the request. \\

\#\#\# Instruction: \\
\textbf{\textcolor{lightblue}{
How many inches are in the circumference of a great circle of a sphere which has a surface area of $196\pi$ sq in? Express your answer in terms of $\pi$.\\
Note: To measure the circumference, take the largest circle on the surface of the sphere.
}} \\

\#\#\# Response: \\
Let’s think step by step. \\
\textbf{\textcolor{lightblue}{
The surface area of a sphere is $4\pi r^2$, so $r^2 = 49$ and $r = 7$. \\       
The circumference of the great circle is $2\pi r = \boxed{14\pi}$. \\
The final answer is: $14\pi$.
}}

}
\end{exmp}

\begin{table*}[htb]
    \centering
    \resizebox{.9\textwidth}{!}{
        \begin{tabular}{cl*{9}{c}}
            \toprule
            Train Ratio & Methods & BoolQ & PIQA & SIQA & HellaS. & WinoG. & ARC-e & ARC-c & OBQA & Avg. \\
            \midrule 
            
            \multirow{3}{*}{5\%} & LoRA & 69.7 & 82.8 & 73.9 & 90.5 & 79.5 & 87.9 & 75.7 & 76.6 & 79.6\\
            & SeLoRA$_{\text{F}}$ & 71.7 & 85.6 & 75.0 & 91.1 & 81.6 & 90.1 & 77.2 & 79.4 & 81.5 \\
            & SeLoRA$_{\text{W}}$ & 71.5 & 86.3 & 75.7 & 91.4 & 81.2 & 90.4 & 76.8 & 78.8 & 81.5 \\

            \midrule
            
            \multirow{3}{*}{10\%} & LoRA & 69.9 & 85.0 & 75.1 & 91.1 & 82.4 & 88.7 & 76.3 & 78.8 & 80.9\\
            & SeLoRA$_{\text{F}}$ & 72.5 & 85.9 & 77.6 & 92.7 & 83.8 & 90.1 & 78.3 & 82.0 & 82.9 \\
            &  SeLoRA$_{\text{W}}$ & 71.4 & 86.0 & 77.8 & 92.6 & 83.8 & 89.3 & 77.3 & 82.4 & 82.6 \\

            \midrule
            
            \multirow{3}{*}{25\%} & LoRA & 72.7 & 84.8 & 76.7 & 92.6 & 85.5 & 88.7 & 76.6 & 82.6 & 82.5 \\
            & SeLoRA$_{\text{F}}$ & 73.9 & 87.7 & 79.5 & 93.9 & 86.4 & 89.1 & 78.2 & 85.4 & 84.3 \\
            & SeLoRA$_{\text{W}}$ & 73.9 & 86.1 & 78.9 & 94.2 & 86.7 & 89.8 & 77.9 & 83.2 & 83.8 \\

            \midrule
            
            \multirow{3}{*}{50\%} & LoRA & 72.9 & 86.7 & 79.1 & 93.6 & 85.3 & 88.7 & 77.5 & 83.4 & 83.4 \\
            & SeLoRA$_{\text{F}}$ & 73.7 & 88.0 & 79.8 & 94.8 & 86.9 & 91.1 & 78.8 & 84.6 & 84.7 \\
            & SeLoRA$_{\text{W}}$ & 73.8 & 87.8 & 80.0 & 95.0 & 86.1 & 90.3 & 79.1 & 85.2 & 84.7 \\

            \midrule
            
            \multirow{3}{*}{100\%} & LoRA & 74.0 & 88.2 & 80.4 & 94.0 & 85.5 & 87.5 & 78.1 & 84.0 & 84.0 \\
            & SeLoRA$_{\text{F}}$ & 74.4 & 89.0 & 81.3 & 95.6 & 87.5 & 90.6 & 80.3 & 87.0 & 85.7 \\
            & SeLoRA$_{\text{W}}$ & 76.0 & 89.3 & 80.6 & 95.9 & 86.7 & 91.0 & 81.4 & 86.6 & 85.9 \\
            
            \bottomrule
        \end{tabular}
    }
    \caption{Full experiment results on LLaMA3$_{\textsc{8b}}$ with various ratios of training data on eight commonsense datasets.
    }
    \label{tab:common_train_ratio}
\end{table*}
\begin{table*}[htb]
    \centering
    \resizebox{.85\textwidth}{!}{
        \begin{tabular}{cl*{9}{c}}
            \toprule
            Rank & Methods & BoolQ & PIQA & SIQA & HellaS. & WinoG. & ARC-e & ARC-c & OBQA & Avg. \\
            \midrule 

            \multirow{3}{*}{16} & LoRA & 73.6 & 87.8 & 80.0 & 93.6 & 85.1 & 87.2 & 77.7 & 83.6 & 83.6 \\
            & SeLoRA$_{\text{F}}$ & 74.8 & 88.9 & 81.1 & 95.3 & 85.3 &  90.1 & 79.3 & 86.4 & 85.2 \\
            & SeLoRA$_{\text{W}}$ & 74.8 & 89.6 & 80.7 & 95.6 & 86.7 & 90.9 & 81.7 & 86.6 & 85.8 \\

            \midrule
            
            \multirow{3}{*}{32} & LoRA & 74.0 & 88.2 & 80.4 & 94.0 & 85.5 & 87.5 & 78.1 & 84.0 & 84.0 \\
            & SeLoRA$_{\text{F}}$ & 74.4 & 89.0 & 81.3 & 95.6 & 87.5 & 90.6 & 80.3 & 87.0 & 85.7 \\
            & SeLoRA$_{\text{W}}$ & 76.0 & 89.3 & 80.6 & 95.9 & 86.7 & 91.0 & 81.4 & 86.6 & 85.9 \\

            \midrule
            
            \multirow{3}{*}{64} & LoRA & 74.4 & 88.8 & 80.3 & 95.1 & 85.4 & 89.0 & 80.0 & 85.2 & 84.8 \\
            & SeLoRA$_{\text{F}}$ & 76.4 & 89.4 & 81.1 & 96.1 & 87.7 & 91.9 & 80.5 & 87.0 & 86.3 \\
            & SeLoRA$_{\text{W}}$ & 76.2 & 89.8 & 80.5 & 96.2 & 87.1 & 92.0 & 80.9 & 87.0 & 86.2 \\

            \midrule
            
            \multirow{3}{*}{128} & LoRA & 76.0 & 88.4 & 80.5 & 95.0 & 86.0 & 90.2 & 80.3 & 86.4 & 85.4 \\
            & SeLoRA$_{\text{F}}$ & 76.6 & 88.8 & 81.5 & 96.1 & 88.2 & 91.6 & 81.4 & 87.6 & 86.5 \\            
            & SeLoRA$_{\text{W}}$ & 76.3 & 89.0 & 80.8 & 96.0 & 86.7 & 91.7 & 81.5 &  87.6 & 86.2 \\

            \midrule
            
            \multirow{3}{*}{256} & LoRA & 76.6 & 88.7 & 80.6 & 96.0 & 86.5 & 91.6 & 80.9 & 87.8 & 86.1 \\
            & SeLoRA$_{\text{F}}$ & 76.2 & 89.6 & 81.1 & 95.9 & 87.4 & 92.1 & 81.1 & 87.4 & 86.4 \\
            & SeLoRA$_{\text{W}}$ & 77.4 & 88.8 & 81.1 & 96.2 & 86.7 & 92.0 & 81.4 & 89.0 & 86.6 \\

            \midrule
            
            \multirow{3}{*}{512} & LoRA & 76.4 & 89.2 & 81.1 & 96.1 & 87.5 & 91.9 & 80.5 & 87.0 & 86.2 \\
            & SeLoRA$_{\text{F}}$ & 76.4 & 89.0 & 81.4 & 96.2 & 87.8 & 92.0 & 81.2 & 87.8 & 86.5 \\
            & SeLoRA$_{\text{W}}$ & 77.0 & 89.2 & 81.0 & 96.3 & 87.2 & 92.1 & 81.8 & 88.2 & 86.6 \\
            
            \bottomrule
        \end{tabular}
    }
    \caption{Full experiment results on LLaMA3$_{\textsc{8b}}$ with various ranks on eight commonsense datasets.
    }
    \label{tab:common_rank}
\end{table*}

\end{document}